\documentclass[journal]{IEEEtran}
\usepackage{amsmath,amsfonts}
\usepackage{algorithmic}
\usepackage{algorithm}
\usepackage{array}
\usepackage{xcolor}
\usepackage{colortbl}  
\usepackage{multirow}
\usepackage{textcomp}
\usepackage{stfloats}
\usepackage{arydshln} 
\usepackage{url}
\usepackage{verbatim}
\usepackage{graphicx}
\usepackage{subcaption}
\usepackage{caption}
\usepackage{booktabs}

\usepackage[pdfborder={0 0 0}]{hyperref}    
\usepackage{cleveref}    
\usepackage{makecell}
\begin{document}

\title{Scale-Aware UAV-to-Satellite Cross-View Geo-Localization: A Semantic Geometric Approach}

\author{Yibin Ye,  Shuo Chen, Kun Wang, Xiaokai Song, Jisheng Dang, Qifeng Yu, Xichao Teng, Zhang Li,~\IEEEmembership{Member,~IEEE}
\thanks{This paper was supported in part by the National Natural Science Foundation of China under Grant 12472189, in part by the Science and Technology Innovation Program of Hunan Province under Grant 2022RC1196. 
Yibin Ye,  Shuo Chen, Kun Wang, Xiaokai Song, Qifeng Yu, Xichao Teng and Zhang Li are with the College of Aerospace Science and Engineering, National University of Defense Technology, Changsha 410073, China, and Hunan Provincial Key Laboratory of Image Measurement and Vision Navigation, Changsha 410073, China. Jisheng Dang is with School of Information Science and Engineering, Lanzhou University, Lanzhou 730000, China. Co-corresponding authors: Zhang Li and Xichao Teng (E-mail: zhangli\_nudt@163.com). 

}

\thanks{Manuscript received April 19, 2021; revised August 16, 2021.}}

\markboth{Journal of \LaTeX\ Class Files,~Vol.~14, No.~8, August~2021}%
{Shell \MakeLowercase{\textit{et al.}}: A Sample Article Using IEEEtran.cls for IEEE Journals}


\maketitle

\begin{abstract} Cross-View Geo-Localization (CVGL) between Unmanned Aerial Vehicle (UAV) images with satellite imagery is crucial for target localization and UAV self-positioning. Although numerous approaches have been proposed in this field, they predominantly rely on the idealized assumption of scale consistency between UAV queries and satellite galleries, neglecting the severe scale ambiguity often encountered in real-world applications. This discrepancy induces significant field-of-view (FOV) misalignment and feature mismatch, drastically reducing CVGL robustness. To bridge this gap, we propose a novel geometric framework that recovers the absolute metric scale from monocular UAV images by leveraging semantic anchors. Specifically, small vehicles (SV) characterized by stable prior scale distributions are identified as optimal metric references due to their ubiquity and high detectability. A Decoupled Stereoscopic Projection Model is proposed to derive the absolute image scale from these semantic targets. By decomposing vehicle dimensions into radial and tangential components, this model rectifies the perspective distortions inherent in 2D detections of 3D stereoscopic SV, thereby enabling precise scale estimation. Furthermore, to mitigate the inherent intra-class size variance of SVs and detection noise, a dual-dimension fusion strategy combined with Interquartile Range (IQR)-based robust aggregation is employed. The estimated global scale serves as a physical constraint to perform scale-adaptive cropping of satellite imagery, facilitating better UAV-to-satellite feature alignment. Extensive experiments on the augmented DenseUAV and UAV-VisLoc datasets demonstrate that our method effectively enhances the robustness of CVGL under unknown UAV image scale conditions. Additionally, the proposed framework exhibits strong applicability in downstream tasks such as passive UAV altitude estimation and 3D model scale recovery. The code and dataset are available at \url{https://github.com/UAV-AVL/U2S-CVGL}\end{abstract}


\section{Introduction}

\begin{figure*}[!h]
    \centering
    \begin{subfigure}{0.59\textwidth}
        \centering
        \includegraphics[page=1, width=\linewidth]{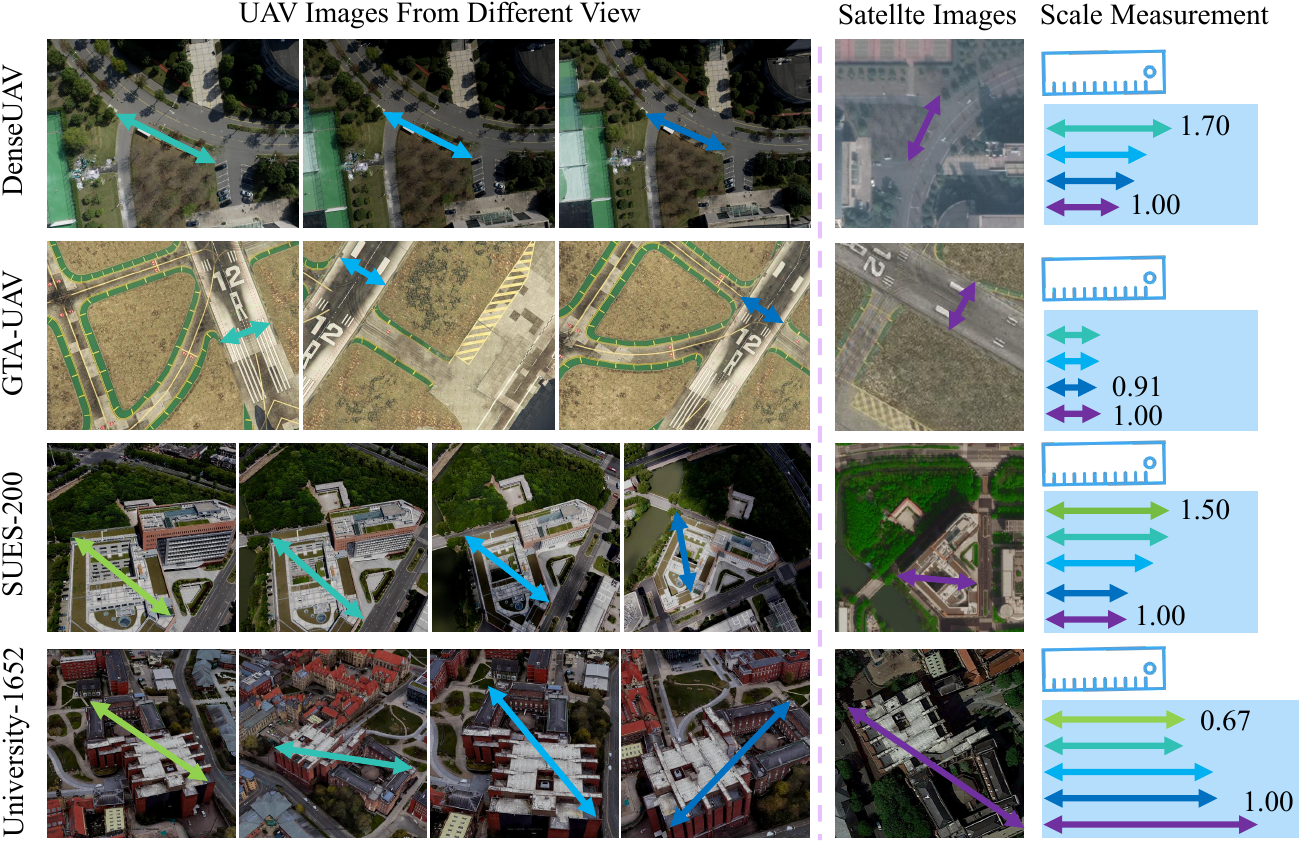} 
        \caption{Scale consistency in existing CVGL datasets}
        \label{subfig:existing_dataset_scale}
    \end{subfigure}
    \hfill 
    \begin{subfigure}{0.38\textwidth}
        \centering
        \includegraphics[page=2, width=\linewidth]{scale1.pdf} 
        \caption{Scale Ambiguity in Real-world Scenarios}
        \label{subfig:unknown_scale_cropping}
    \end{subfigure}
    \caption{Scale comparison in UAV-to-Satellite CVGL: (a) For existing datasets (DenseUAV, GTA-UAV, SUES-200, University-1652), UAV image scales (marked by arrows) are consistent with satellite images (purple arrow, normalized to 1.0), with maximum difference  not exceeding a factor of 2; (b) Under unknown scale, imprecise satellite cropping causes huge scale/FOV discrepancies between UAV images and satellite crops.}
    \label{fig:cvgl_scale_comparison}
\end{figure*}

\IEEEPARstart{W}{ith} the extensive deployment of Unmanned Aerial Vehicles (UAVs) in remote sensing and urban monitoring, UAV-to-Satellite Cross-View Geo-Localization (CVGL) has attracted increasing attention \cite{LPN, FSRA, shen2023mccg, DAC, TirSA}. This task is generally modeled as an image retrieval problem, which aims to locate a UAV by matching its aerial view (query) to the most similar candidate satellite image (gallery) \cite{he2023foundloc}. Such technology is important for target localization and UAV self-positioning in GNSS (Global Navigation Satellite System)-denied environments \cite{DenseUAV}. Although numerous studies have been conducted on benchmarks like University-1652 \cite{University-1652}, SUES-200 \cite{sues200}, DenseUAV \cite{DenseUAV} and GTA-UAV \cite{ji2024game4loc}, these datasets often rely on an idealized assumption: the UAV images are pre-rectified to a scale approximating that of the satellite image.  As illustrated in Fig. \ref{fig:cvgl_scale_comparison}a, the scale difference between UAV images and corresponding satellite images generally does not exceed a factor of 2. This scale consistency enhances the feature similarity of corresponding regions, making retrieval relatively straightforward under such ideal conditions. However, real-world scenarios often suffer from scale ambiguity. When the absolute scale of the UAV image is inaccurate or even unknown, it becomes infeasible to precisely crop the corresponding satellite imagery. Consequently, although the UAV query and the satellite gallery are resized to identical pixel resolutions for image retrieval network input, they exhibit significant discrepancies in both Field-of-View (FOV) and physical scale. Specifically, FOV misalignment introduces substantial redundant background information or results in context loss, while scale inconsistency hinders the alignment of local semantic features. These coupled discrepancies inevitably expand the search space and degrade feature matching accuracy, thereby reducing retrieval robustness \cite{DenseUAV, ye2025exploring, zhang2025GeoRVLF}.

In practical applications, the scale (or Ground Sample Distance, GSD) of UAV image is typically estimated using camera pose and intrinsics. However, in GNSS-denied environments where flight positioning is interfered with \cite{GPS_jamming}, obtaining accurate pose data is difficult, which hinders image scale estimation and subsequent CVGL task. Moreover, for intelligence analysis based on UAV images from social media platforms, metadata such as flight altitude is often incomplete, which invalidates standard scale estimation approaches relying on explicit imaging geometry. Other absolute scale estimation methods also face various limitations. For instance, Structure-from-Motion (SfM) methods can recover scale but requires multi-view observations. This does not work for a single-frame UAV image. Although some onboard sensors like Inertial Navigation Systems (INS), barometers, LiDARs and depth cameras can provide pose or distance information to assist scale estimation, they have drawbacks. Barometers only determine absolute altitude and cannot provide relative altitude above ground without external Digital Surface Models (DSMs). Although the INS can estimate the UAV’s pose relative to its starting point, they do not directly provide the absolute distance from the camera to ground scenes in the current frame. Furthermore, low-cost MEMS-based INS suffer from significant drift \cite{he2024leveraging}, and LiDARs and depth cameras are limited by sensing range. Recently, deep learning-based monocular metric depth estimation models have shown promise in indoor and autonomous driving tasks but suffer from domain gaps when applied to aerial views that differ from their ground-level training data \cite{yang2024depth}.  Consequently, recovering the absolute scale of a monocular UAV image to assist CVGL tasks remains challenging in these constrained scenarios. Beyond methodological challenges, research in this domain is further constrained by the limitations of existing benchmarks. For instance, datasets like DenseUAV \cite{DenseUAV} lack continuous satellite image, providing only fixed crops that preclude the validation of scale-adaptive cropping strategies. Similarly, datasets such as UAV-VisLoc \cite{UAV-VisLoc} record only absolute altitude, lacking the relative altitude essential for accurate scale estimation. Therefore, to enhance the robustness of the UAV-to-Satellite CVGL task, it is necessary to investigate more effective methods for monocular absolute scale estimation and establish suitable benchmarks for validation.


In this work, we take a different perspective on the absolute scale recovery problem. Instead of explicitly estimating camera altitude or scene depth, we exploit semantic objects with stable physical geometry as anchors to recover image scale. We observe that many man-made objects in aerial imagery exhibit relatively consistent real-world dimensions. Among them, small vehicles uniquely satisfy three desirable properties: (i) low variance in physical size, (ii) ubiquitous presence in urban and suburban environments, and (iii) high detectability using modern object detectors. These properties make small vehicles particularly suitable for serving as reliable scale references across diverse scenes. Building on this insight, we propose a semantic geometric framework that leverages small vehicle detections to recover absolute scale, enabling Scale-Aware UAV-to-Satellite Cross-View Geo-Localization.

Although object detection for scale estimation is a well-established concept, our approach is distinguished by its specific focus and robustness. Unlike existing measurement methods that rely on cooperative targets with fixed scales (like markers) \cite{mraz2020using, yoon20213d}, our approach works with non-cooperative targets characterized by inherent intra-class size variance. In addition, our application primarily focuses on coarse scale recovery problem for CVGL, which differs significantly from other high-precision photogrammetry works. To address the stereoscopic effects of 3D objects and intra-class variance, we propose a Decoupled Stereoscopic Projection Model combined with a Dual-Dimension Scale Recovery strategy. To support research addressing scale ambiguity in CVGL, we provide augmented extensions for existing benchmarks, incorporating accurate relative scale annotations for the UAV-VisLoc dataset and spatially continuous satellite imagery for the DenseUAV dataset. Extensive experiments demonstrate that our method can reliably estimate absolute scale with relative  error $\leq 10\%$ when sufficient detectable targets are present, thereby significantly improving the robustness of CVGL methods. Additionally, our method can be applied to UAV altitude estimation under GNSS-denied conditions and scale recovery for dimensionless 3D reconstruction results.

In summary, the main contributions of this paper are as follows:
\begin{enumerate}
    \item We comprehensively analyze the impact of scale discrepancy on UAV-to-Satellite CVGL task, highlighting scale consistency as a critical yet often overlooked determinant for real-world robustness.
    
    \item We propose a robust geometric framework that leverages semantic anchors to estimate absolute scale from monocular UAV images, addressing challenges posed by perspective distortion and intra-class dimension variance.
    
    \item Extensive experiments demonstrate that our method significantly enhances CVGL robustness under scale ambiguity, while also serving as a versatile solution for UAV altitude estimation and 3D model scale recovery.
    
    \item We release augmented extensions for the UAV-VisLoc and DenseUAV datasets, incorporating both scale annotations and geographically continuous satellite galleries to facilitate future research in monocular scale estimation and downstream applications.
\end{enumerate}

\section{Related Work}
\label{sec:related_work}

\subsection{UAV-to-Satellite Cross-View Geo-Localization}
The research landscape of UAV-to-satellite CVGL has witnessed a significant paradigm shift over the past decade. Early methodologies primarily relied on hand-crafted descriptors aggregated via techniques like Vector of Locally Aggregated Descriptors (VLAD) \cite{VLAD} or Bag-of-Words (BoW) \cite{BOW} to match visual contents. Driven by the superior feature representation of neural networks, contemporary methods aim to bridge the substantial geometric and visual gaps between aerial and satellite views \cite{LPN, zhu2022transgeo, shen2023mccg, wu2024camp}. However, most research relies on benchmarks like University-1652 \cite{University-1652, LPN, shen2023mccg, ye2024coarse}, SUES-200 \cite{sues200}, DenseUAV \cite{DenseUAV} and  GTA-UAV \cite{ji2024game4loc}. These datasets operate under idealized assumptions where UAV images are pre-rectified to match the satellite images' scale. Although some recent works have attempted to address the scale discrepancy found in real-world scenarios, limitations remain. Dai et al. \cite{DenseUAV} and Ye et al. \cite{ye2025exploring} both preliminarily analyzed the impact of scale discrepancy but lacked an in-depth discussion on mitigation strategies. To improve CVGL robustness under scale ambiguity, He et al. \cite{he2024leveraging} warped UAV images to multiple scales to verify retrieval consistency. Although this brute-force strategy boosts accuracy, it compromises computational efficiency and requires heuristic consistency thresholds. Zhang et al. \cite{zhang2025GeoRVLF} proposed a multi-scale pyramid search strategy combined with image matching inliers for verification. While this reduces search time compared to brute-force methods, it needs auxiliary local feature matching algorithms. Alternatively, some approaches leverage camera pose priors to rectify UAV images to a scale and viewpoint consistent with satellite image \cite{LSVL, sui2022fast}. While these pose-guided methods improve accuracy, they rely on reliable GNSS signals or onboard sensor data. Unlike methods that rely on computationally expensive multi-scale inputs or external sensor data, our work utilizes semantic object priors to recover the absolute scale directly from the image content, serving as a plug-and-play module to enhance CVGL robustness.

\subsection{Absolute Scale Recovery for UAV Images}

In monocular geometric reasoning, absolute scale, metric depth, and the 6 DoF camera pose are intrinsically coupled. Recovering one attribute effectively constrains the others. Consequently, approaches for addressing the scale ambiguity problem can be categorized into photogrammetric and sensor-based, learning-based, and object-based methods.

\textbf{Photogrammetric and Sensor-based Scale Recovery.} 
Traditional multi-view techniques, such as Structure-from-Motion (SfM), reconstruct 3D structure based on geometric constraints. However, they strictly require multi-view observations, known camera intrinsics and initial pose priors. Such prerequisites are impossible to meet for single-frame UAV images or images with stripped metadata. While onboard sensors (e.g., Barometers, IMUs) can provide direct metric measurements, they are prone to hardware drift and often fail to provide the relative altitude above ground required for accurate scale estimation. Moreover, active sensors like LiDAR and depth cameras are constrained by limited measurement ranges and high hardware costs, making them less viable for consumer UAVs operating at varying altitudes.

\textbf{Learning-based Scale Recovery.} In the absence of geometric constraints, Monocular Depth Estimation (MDE) models \cite{hu2024metric3d, yang2024depth, depth_anything_v2} utilize deep neural networks to infer depth from visual cues. Despite their success in ground-level scenarios, these models suffer from significant domain gaps when applied to UAV images \cite{florea2025tandepth, alaaeldin2025co}. Crucially, most general-purpose MDE methods yield only relative depth maps \cite{Ranftl2022, depth_anything_v1}, which can not be used for absolute scale recovery. Recently, end-to-end geometric foundation models like DUSt3R \cite{dust3r_cvpr24}, VGGT \cite{wang2025vggt}, and MUSt3R \cite{must3r_cvpr25} have achieved breakthrough performance in 3D attributes prediction. However, DUSt3R inherently yields scale-free structures, while VGGT and MUSt3R, despite their capability for metric estimation, primarily rely on multi-view consistency or data-driven priors, leading to potential generalization risks when recovering absolute scale in unseen monocular aerial domains \cite{wu2025evaluation}. 

\textbf{Object-based Scale Recovery.} Distinct from implicit deep learning methods, object-based approaches explicitly leverage the physical dimensions of known targets as metric anchors to solve the scale ambiguity. Some research utilized cooperative targets with fixed patterns, such as helipads \cite{mondragon20103d} or markers \cite{yoon20213d}, to recover the camera pose along with image scale. However, these methods are restricted to controlled environments and cannot adapt to the dynamic, unstructured scenes encountered during UAV exploration. To address open-world challenges, attention has shifted to utilizing ubiquitous non-cooperative objects. Although some studies leverage vehicles, they primarily target different objectives, such as utilizing vehicle bounding boxes to estimate inter-object distances for autonomous driving \cite{rezaei2015robust, liu2022vehicle, raj2025distance} or to estimate drone velocity \cite{nam2020moving}, rather than recovering the global image scale. In the specific domain of monocular absolute scale recovery, leveraging known semantic priors to resolve scale ambiguity is a well-established practice. For instance, Dubsk{\'a} et al. \cite{dubska2014fully} demonstrated that the statistical distribution of vehicle sizes can fully calibrate roadside surveillance cameras, and CubeSLAM \cite{yang2019cubeslam} jointly optimizes camera trajectory and object cuboids to recover the absolute scale of the map. While these studies validate the utility of semantic anchors for scale recovery, they primarily focus on ground-level scenarios, which present a viewing geometry vastly different from UAV aerial imaging. Consequently, our work focuses on utilizing the 2D bounding boxes of 3D small vehicles to recover the image scale.

\section{Scale-aware Cross-view Geo-Localization}

In this section, we first elaborate on the semantic anchor selection strategy and the instance-level dual-dimension scale estimation model in Sec. \ref{subsec:anchor_selection} and Sec. \ref{subsec:dual_dimension}, respectively. Subsequently, we present the robust global aggregation pipeline in Sec. \ref{subsec:global_recovery} and introduce its application to UAV-to-Satellite  CVGL in Sec. \ref{subsec:scale_cvgl}.

\subsection{Semantic Anchor Selection Strategy}
\label{subsec:anchor_selection}
Identifying an optimal metric anchor from the UAV view is the prerequisite for scale recovery. Due to the cross-scene flight characteristics of UAVs, this selection is non-trivial. While natural environments lack consistent geometric priors, man-made urban environments contain various standardized objects (e.g., basketball courts, road markings). However, determining which object serves as the optimal anchor requires careful consideration. We posit that an ideal semantic anchor must satisfy three criteria: (1) \textbf{Ubiquity}: The object must appear frequently across diverse scenes and varying UAV altitudes to ensure availability during flight; (2) \textbf{Geometric Stability}: The object should possess an approximate standard scale with minimal intra-class variance in its physical dimensions;and (3) \textbf{Detectability}: It must be detectable with high precision using off-the-shelf object detection algorithms.
\begin{figure*}[!t]
    \centering
    \includegraphics[width=1.0\linewidth]{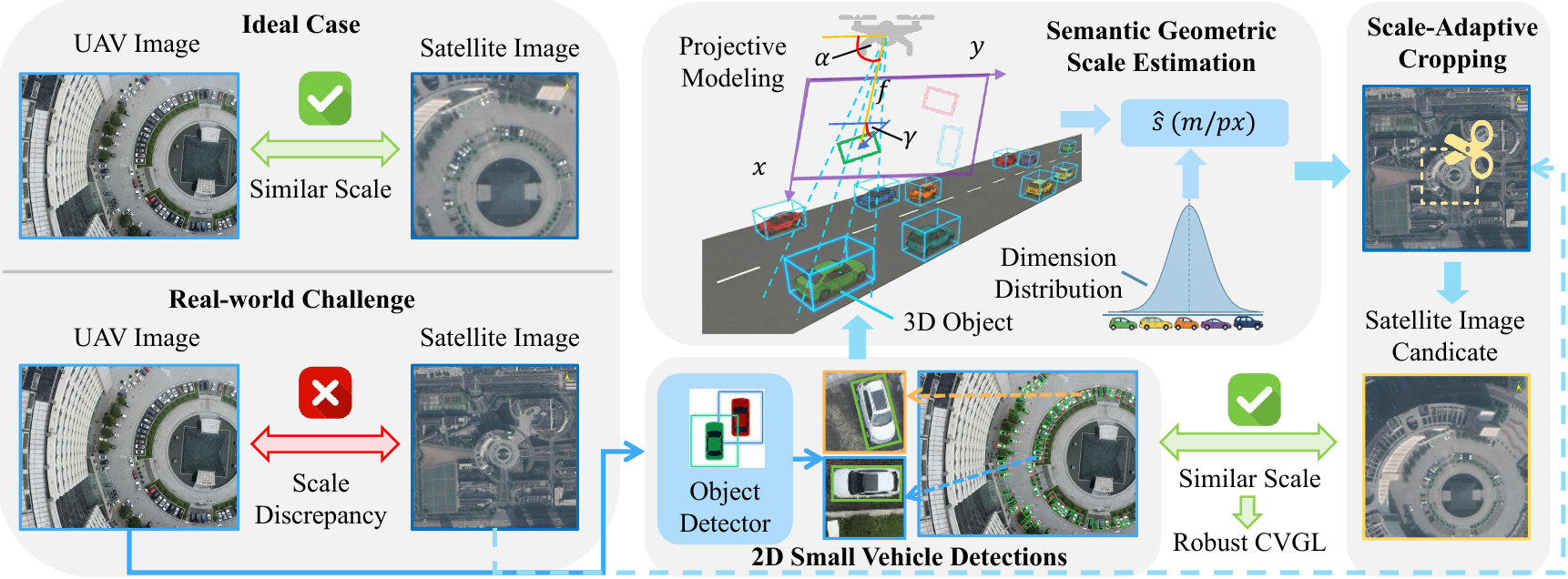} 
    \caption{Overview of the proposed Scale-aware CVGL framework. The method first utilizes oriented bounding boxes of small vehicles as semantic anchors. Notably, off-center targets (orange frame) exhibit significant stereoscopic effects, necessitating the incorporation of vehicle height into the modeling. To address this, we employ a decoupled stereoscopic projection model combined with statistical dimension distributions to estimate the single-instance absolute scale. Finally, robust global scale estimation is conducted to guide the scale-adaptive cropping of satellite imagery, ensuring robust CVGL.}
    \label{fig:scale_method}
\end{figure*}

To establish reliable metric anchors for UAV image scale recovery, we utilize the DOTA-v2.0 dataset \cite{dota_v2}  for comprehensive analysis. DOTA-v2.0 is a popular object detection benchmark, particularly valuable for our study as it exclusively comprises remote sensing images acquired from satellite and aerial platforms, ensuring domain consistency with the UAV imagery we focus on. The dataset contains 11,268 images and approximately 1.8 million object instances across 18 categories, exhibiting significant diversity in spatial resolutions, scene types, and object distributions. This rich diversity facilitates thorough statistical analysis of potential metric anchors under various real-world conditions. 

\begin{figure}[!h]
    \centering
    
    \begin{subfigure}{0.48\textwidth}
        \centering
        \includegraphics[page=1, width=\linewidth]{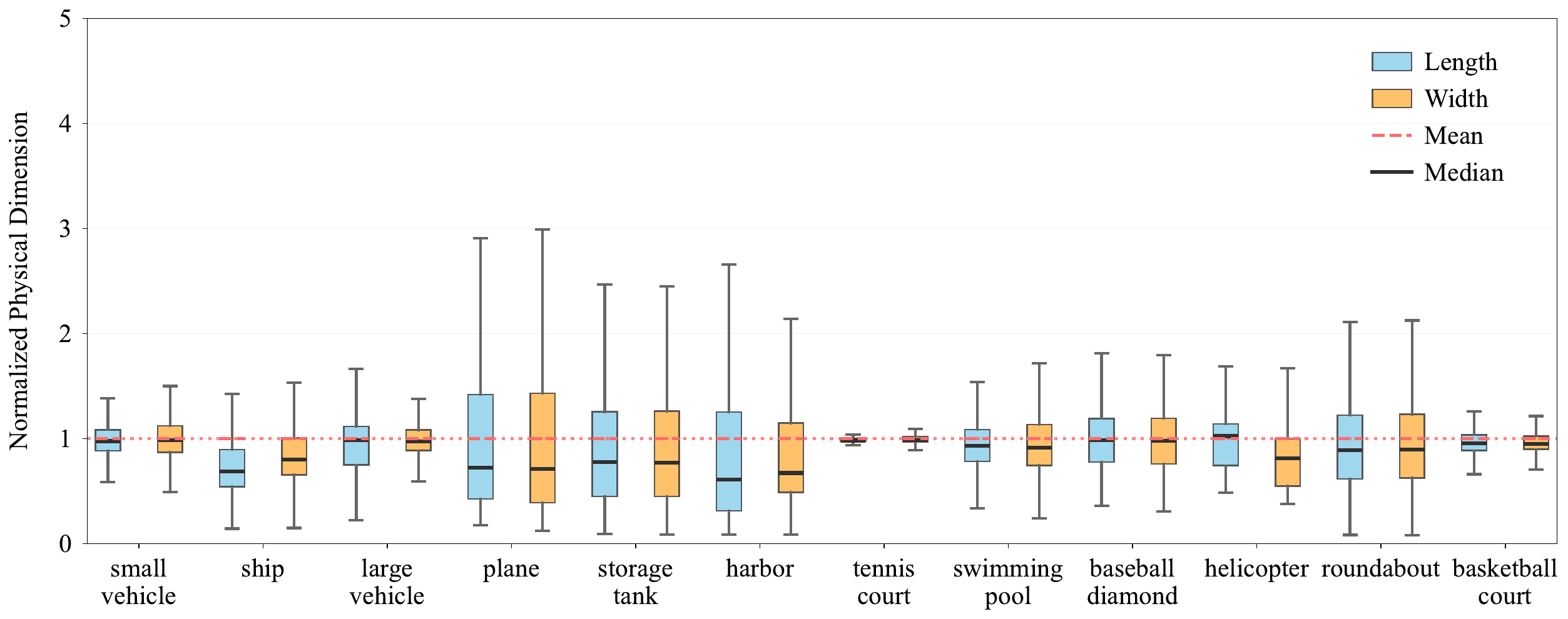}
        \caption{Dimension distribution}
    \end{subfigure}
    \vfill
    \begin{subfigure}{0.48\textwidth}
        \centering
        \includegraphics[page=2, width=\linewidth]{dota_normalized_analysis.pdf}
        \caption{Category frequency}
    \end{subfigure}
    
    \caption{DOTA dataset object category analysis. (a) Normalized length (blue) and width (orange) distributions. (b) Average instances per image frequency.}
    \label{fig:dota_analysis_short}
\end{figure}

To statistically evaluate the ubiquity and geometric stability of each object category, we calculated the physical dimensions variation of different categories as well as the distribution of instances per image. The statistic in Fig. \ref{fig:dota_analysis_short} reveals distinct patterns among various object categories. Large structures such as tennis courts and basketball court exhibit low dimensional variance but suffer from sparse distribution, making them unavailable in most random frames. Conversely, large vehicles (e.g., trucks, buses) demonstrate higher distribution frequency but suffer from significant dimensional variance, particularly in length direction. \textbf{Small Vehicles (SV)} optimally satisfy all three selection criteria: they are densely distributed throughout the dataset, their physical dimensions follow a compact distribution pattern, and they achieve detection accuracy generally above 75\% on the DOTA-v2.0 dataset using state-of-the-art methods \cite{wang2025oriented}. The consistent performance across different detection methodologies confirms their reliability as metric references.
Therefore, we select small vehicles as our primary metric anchors for UAV image  scale recovery.

\subsection{Dual-Dimension Scale Recovery}
\label{subsec:dual_dimension}
\textbf{Problem Formulation.}
Given the camera intrinsic matrix $\mathbf{K}$ and the 2D corner coordinates $\mathcal{P}$ of a detected vehicle OBB, our goal is to estimate the instance-level metric scale $s$ (meters per pixel). The intrinsic matrix $\mathbf{K}$ is obtained from camera calibration and is written as
\begin{equation}
\mathbf{K}=
\begin{bmatrix}
f_x & 0 & c_x\\
0 & f_y & c_y\\
0 & 0 & 1
\end{bmatrix},
\end{equation}
where $(c_x, c_y)$ is the principal point and $(f_x, f_y)$ are focal lengths in pixel units. Since $f_x$ and $f_y$ are typically similar, we use the effective focal length $f=(f_x+f_y)/2$. The camera pitch angle $\theta$ is used for scale estimation. In many practical deployments, $\theta$ is approximately constant due to gimbal stabilization and can be set to the nominal mounting angle (e.g., $\theta=-90^\circ$ for nadir-looking flight); when unavailable, we use a reasonable default and treat residual deviations as noise. The roll angle is assumed to be zero, as it can be actively stabilized by the gimbal.

\textbf{Modeling Ambiguity.}
In low-altitude UAV imagery, small vehicles manifest as typical 3D stereoscopic targets rather than planar objects. As illustrated in Fig. \ref{fig:scale_method}, targets located away from the image center (e.g., the orange frame in the zoomed-in view) exhibit significant stereoscopic effects compared to near-center ones (blue frame). This observation underscores the necessity of incorporating vehicle height into the geometric modeling. Although Oriented Bounding Box (OBB) detectors provide directional cues, the predicted box essentially represents the 2D enclosing rectangle of the vehicle's 3D volumetric projection under perspective distortion. Since reliance is placed solely on the ambiguous corner coordinates of this bounding box without specific semantic keypoints (e.g., wheels or windshield corners), recovering the precise 3D pose or structure using standard Perspective-n-Point (PnP) approaches is infeasible. To resolve this ambiguity, we propose a decoupled stereoscopic projection model. By assuming the small vehicle is a rigid body resting on the ground plane, the   statistical priors of both length and width are utilized to mathematically decouple the perspective effects and robustly recover the scale.

\textbf{Viewing Angle Calculation.}
The perspective distortion of a vehicle is strictly determined by its viewing angle relative to the ground. Let the UAV camera's optical center be the origin. Given the pixel coordinate $(u, v)$ of the vehicle center and the focal length $f$, the viewing ray vector is $\mathbf{v}_{ray} = [u-c_x, v-c_y, f]^\top$. Since the camera's roll angle is set to zero, the ground normal vector $\mathbf{n}_{up}$ in the camera coordinate system is derived using the pitch angle $\theta$:
\begin{equation}
    \mathbf{n}_{up} = [0, -\cos\theta, -\sin\theta]^\top
\end{equation}.
The viewing elevation angle $\alpha$ (where $\alpha=90^\circ$ denotes a nadir view) is computed via the geometric dot product:
\begin{equation}
    \sin\alpha = \frac{|\mathbf{v}_{ray} \cdot \mathbf{n}_{up}|}{\|\mathbf{v}_{ray}\|_2}.
    \label{equ:alpha}
\end{equation}

\textbf{Relative Orientation Derivation.}
To accurately model the stereoscopic projection, the relative rotation angle $\gamma$ between the vehicle's orientation and the camera's radial direction must be determined. The radial vector $\mathbf{v}_{rad}$ is defined from the principal point $(c_x, c_y)$ to the vehicle center $(u, v)$ on the image plane. The vehicle's orientation vector $\mathbf{v}_{edge}$ is derived from the long edge of the detected OBB. Here, we assume the detected OBB's long side corresponds to the vehicle's longitudinal axis, which holds true for most aerial viewpoints provided $\alpha$ is not extremely small. The angle $\gamma$ is then computed as:
\begin{equation}
    \mathbf{v}_{rad} = [u-c_x, v-c_y]^\top, \quad \cos\gamma = \frac{|\mathbf{v}_{rad} \cdot \mathbf{v}_{edge}|}{\|\mathbf{v}_{rad}\|_2 \|\mathbf{v}_{edge}\|_2}.
    \label{equ:gamma}
\end{equation}
This geometric constraint determines whether the vehicle is observed primarily from the side (radial) or the front/rear (tangential).

\textbf{Decoupled Projection Model.}
The 3D vehicle dimensions are decomposed into radial (along the viewing direction) and tangential components to model the "stereoscopic inflation" effect. Taking the vehicle length $L_{car}$ as an example, its effective projected physical size $L_{eff}$ is derived from $\alpha$ and $\gamma$:
\begin{equation}
    T_{rad}^L = (L_{car}\sin\alpha + H_{car}\cos\alpha)\cos\gamma,
\end{equation}
\begin{equation}
    T_{tan}^L = L_{car}\sin\gamma,
\end{equation}
\begin{equation}
    L_{eff} = \sqrt{(T_{rad}^L)^2 + (T_{tan}^L)^2}.
\end{equation}
Here, the radial component $T_{rad}^L$ accounts for the perspective foreshortening and the visible height $H_{car}$, while the tangential component $T_{tan}^L$ remains distortion-free. The same logic is applied to the vehicle width $W_{car}$ to obtain $W_{eff}$.

\textbf{Instance Scale Recovery.}
The goal is to recover the nadir-equivalent scale $s$ (meters per pixel) for each instance. Two scale candidates are calculated based on the detected pixel length $l_{pix}$ and width $w_{pix}$. The $\sin\alpha$ term normalizes the slant range effect, converting the oblique observation to a consistent vertical scale:
\begin{equation}
    s_{len} = \frac{L_{eff} \cdot \sin\alpha}{l_{pix}}, \quad s_{wid} = \frac{W_{eff} \cdot \sin\alpha}{w_{pix}}.
\end{equation}
To improve robustness, the two candidates are fused by simple equal weighting and the final instance scale is computed as:
\begin{equation}
s_{i} = \frac{s_{len} + s_{wid}}{2}.
\end{equation}

\subsection{Robust Global Scale Recovery}
\label{subsec:global_recovery}
Individual scale estimates $s_i$ inevitably contain noise due to multiple factors. First, potential model misdetections can introduce false positives. Second, visual occlusions may result in incomplete bounding boxes, distorting the geometric attributes required for accurate scale recovery. Third, inherent non-uniformity in vehicle dimensions, such as targets with extreme aspect ratios or sizes deviating significantly from the statistical mean can lead to outliers. To mitigate these uncertainties, a robust aggregation pipeline is employed to determine the global image scale $\hat{s}$.

\textbf{Reliability Filtering.}
Before aggregation, low-quality detections are filtered to ensure data reliability. Let $\mathcal{D} = \{d_1, ..., d_M\}$ be the initial set of detections. We use the detection confidence threshold $\tau_{conf}$ to filter out false positives, where $\text{score}(\cdot)$ denotes the confidence score output by the object detector, defining the valid detection subset $\mathcal{D}_{valid}$ as:
\begin{equation}
    \mathcal{D}_{valid} = \{d_k \in \mathcal{D} \mid \text{score}(d_k) > \tau_{conf}\}.
\end{equation}
Subsequently, to avoid statistical bias caused by sparse sampling, scale estimation proceeds only if the number of valid semantic anchors meets a minimum count threshold $N_{min}$:
\begin{equation}
    |\mathcal{D}_{valid}| \ge N_{min}.
\end{equation}

\textbf{IQR-based Aggregation.}
The Interquartile Range (IQR) method is adopted to robustly filter statistical outliers from the valid estimates. Let $\mathcal{S} = \{s_1, ..., s_N\}$ be the set of valid scales derived from $\mathcal{D}_{valid}$. The first and third quartiles ($Q_1, Q_3$) are calculated, and the inlier range is defined as $\Omega = [Q_1 - 1.5 \cdot IQR, \ Q_3 + 1.5 \cdot IQR]$.
The final global scale $\hat{s}$ is computed as the mean of the inliers:
\begin{equation}
    \hat{s} = \frac{1}{|\mathcal{S}_{in}|} \sum_{s \in \mathcal{S}_{in}} s, \quad \text{where } \mathcal{S}_{in} = \{s \in \mathcal{S} \mid s \in \Omega\}.
\end{equation}
\subsection{Scale-aware CVGL}
\label{subsec:scale_cvgl}
The estimated global scale $\hat{s}$ serves as a critical geometric constraint for downstream CVGL.

\textbf{Average Image Scale Recovery.}
To match the UAV image with the satellite imagery, the average spatial resolution (GSD) of the UAV image must be determined to perform scale-adaptive cropping. Using the estimated global scale $\hat{s}$, we derive the relative flight altitude $H_{uav} = \hat{s} \cdot f$ as an intermediate physical variable. Accounting for the camera pitch angle $\theta$ and equivalent focal length $f$, the average spatial resolution $r$ is calculated as:
\begin{equation}
    r = \frac{H_{uav}}{|\sin\theta|} \cdot \frac{1}{f} = \frac{\hat{s}}{|\sin\theta|}.
    \label{equ:scale_estimate}
\end{equation}
This formula effectively adapts the nadir-equivalent scale $\hat{s}$ to the oblique viewing geometry of the UAV.

\textbf{Scale Alignment.}
Let the satellite imagery have a fixed resolution $GSD_{sat}$. The metric footprint $L_{met}$ of the UAV's image width $W_{img}$ is computed using the estimated resolution $r$, and the corresponding crop size $S_{sat}$ on the satellite imagery is determined:
\begin{equation}
    L_{met} = r \cdot W_{img}, \quad S_{sat} = \frac{L_{met}}{GSD_{sat}}.
\end{equation}
This ensures scale consistency between the query UAV image and the satellite gallery, significantly improving retrieval accuracy under varying flight altitudes.

\begin{algorithm}[t]
\caption{Dual-Dimension Scale Recovery}
\label{alg:scale_recovery}
\begin{algorithmic}[1]
\REQUIRE Image $I$, focal length in pixels $f$, Pitch $\theta$, Priors $\{L_{car}, W_{car}, H_{car}\}$, Sat. GSD $GSD_{sat}$
\ENSURE Valid scales $\mathcal{S}_{valid}$

\STATE \textbf{Initialize:} $\mathcal{S}_{valid} \leftarrow \emptyset$

\FOR{each detected vehicle instance $i$ in image $I$}
    \STATE Obtain bounding box $B_i$ and center pixel $(u_i, v_i)$
    \STATE Compute viewing angles $(\alpha, \gamma)$ from $(u_i,v_i)$ and $f$
    \STATE Compute $h$ using $\theta$ and $H_{car}$ \COMMENT{height difference along optical axis}
    \STATE Compute $L_{eff}$ and $W_{eff}$ using $(\alpha,\gamma,h)$ and $(L_{car},W_{car})$
    \STATE Compute length-based scale $s_{len} \leftarrow L_{eff}/l_{img}$
    \STATE Compute width-based scale $s_{wid} \leftarrow W_{eff}/w_{img}$
    \STATE Fuse by equal weighting: $s_i \leftarrow (s_{len} + s_{wid})/2$
    \STATE Append $s_i$ to $\mathcal{S}_{valid}$
\ENDFOR

\STATE $\hat{s} \leftarrow \mathrm{mean}(\mathcal{S}_{valid})$
\STATE $r \leftarrow \hat{s}/|\sin\theta|$ \COMMENT{Average Spatial Resolution (m/px), remove sign ambiguity}
\STATE $GSD \leftarrow r \cdot GSD_{sat}$ \COMMENT{Ground Sampling Distance}

\RETURN $\mathcal{S}_{valid}$
\end{algorithmic}
\end{algorithm}

\section{Experiments}
\label{sec:experiments}

In this section, we evaluate the proposed framework from three perspectives: 
(1) The effectiveness of scale-adaptive matching in Cross-View Geo-Localization (CVGL), which is our primary contribution; 
(2) The accuracy of the UAV flight altitude estimation; 
(3) The practical application in recovering the absolute metric scale for 3D reconstruction orthophotos.

\subsection{Scale-Adaptive Cross-View Geo-Localization}
\label{subsec:exp_cvgl}

This section assesses how the proposed scale estimation method improves geo-localization performance by resolving the scale discrepancy between UAV and satellite imagery.
\begin{table*}[!b]
    \centering
    \caption{Comparison of dataset characteristics. $f$ denotes the equivalent focal length in pixels.}
    \label{tab:dataset_specs}
    \resizebox{\linewidth}{!}{
    \begin{tabular}{l c c c c c c l}
        \toprule
        \textbf{Dataset} & \textbf{Relative Altitude} & \textbf{Viewpoint} & \textbf{Scenes} & \textbf{Platform} & \textbf{UAV Images} & \textbf{$f$} & \textbf{Main Limitations} \\
        \midrule
        DenseUAV \cite{DenseUAV} & 80m/90m/100m & Nadir & Urban (Campus) & Multi-rotor & 9,099 & $\approx$ 1000 px & Fixed satellite crops \\
        UAV-VisLoc \cite{UAV-VisLoc} & 300-600m & Near-Nadir & Urban/Town/Wild & Fixed Wing/Multi-rotor & 6,742 & $\approx$ 4000 px & Absolute altitude only; Noisy poses;\\
        \bottomrule
    \end{tabular}
    }
\end{table*}
\begin{figure*}[!b]
    \centering
    \includegraphics[width=1.0\linewidth]{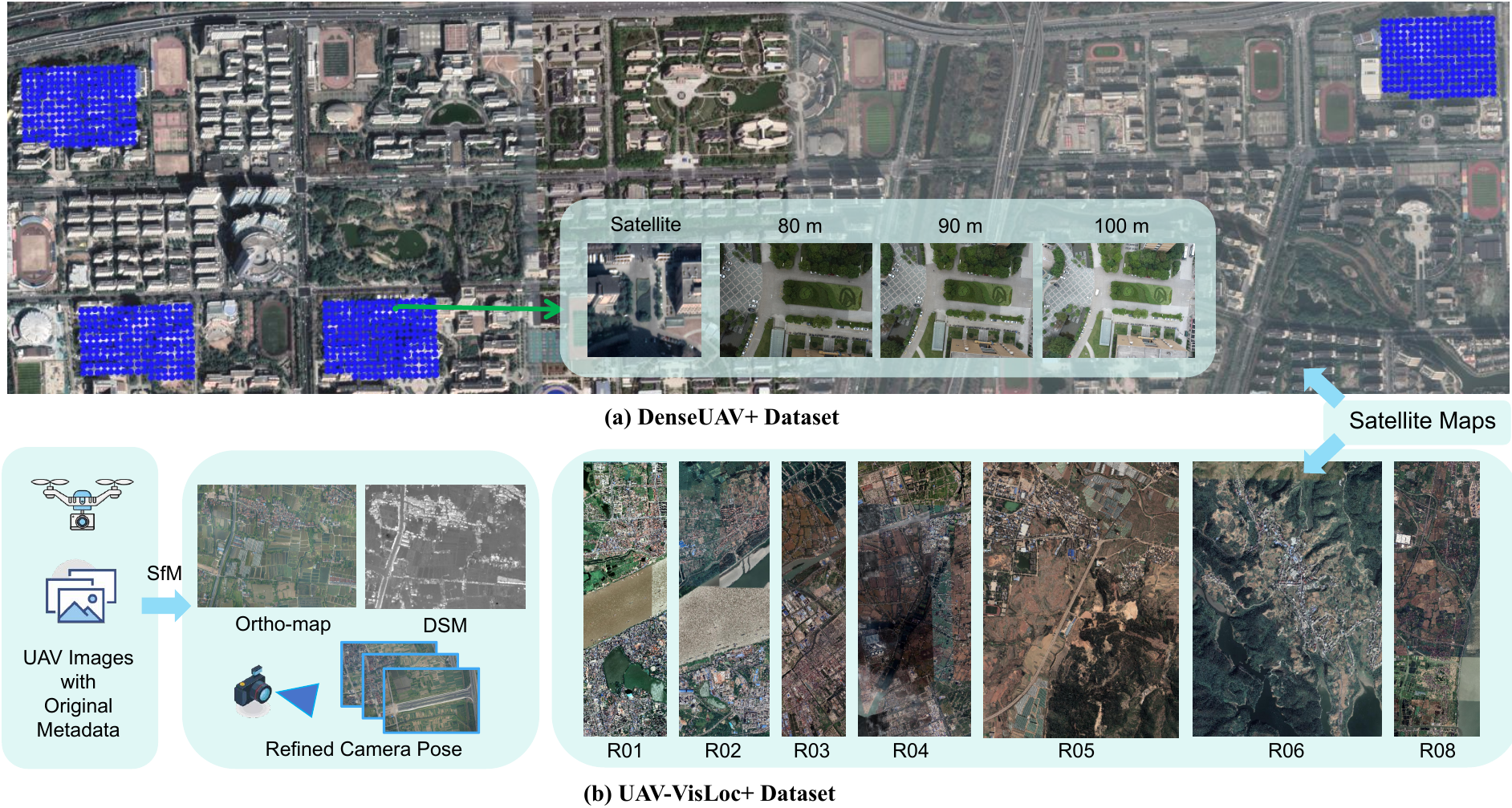}
    \caption{
    Overview of the refined datasets.
    (a) DenseUAV+: expanded satellite maps enabling scale-adaptive cropping at different altitudes. The blue dots denote the UAV trajectories.
    (b) UAV-VisLoc+: SfM-based refinement producing ortho-map/DSM and refined poses for accurate relative altitude.}
    \label{fig:dataset_overview}
\end{figure*}
\subsubsection{Dataset Construction and Augmentation}
\label{subsubsec:dataset_construction}

To rigorously evaluate the proposed method, we construct two augmented benchmarks: \textbf{DenseUAV+} and \textbf{UAV-VisLoc+}. These are derived from the original DenseUAV \cite{DenseUAV} and UAV-VisLoc \cite{UAV-VisLoc} datasets but have been significantly modified to meet the experimental requirements of absolute scale estimation. It is worth noting that while other prominent benchmarks such as University-1652 \cite{University-1652} and SUES-200 \cite{sues200} are widely utilized in the CVGL community, they  do not provide accessible camera intrinsics or precise relative altitude ground truth, rendering them incompatible with our experimental requirement.

\textbf{Overview of Original Datasets.}
The characteristics of the original datasets are summarized in Table \ref{tab:dataset_specs}. 
 DenseUAV \cite{DenseUAV} focuses on low-altitude flights (80m--100m) over urban university campuses using multi-rotor drones (DJI Phantom 4 RTK), offering high-overlap nadir imagery. 
 In contrast, UAV-VisLoc \cite{UAV-VisLoc} spans a significantly broader altitude relative range (300 m--600 m) and diverse scenes (e.g., urban, towns, wild)  over  11 geographically distinct regions (R01–R11), utilizing both fixed-wing and multi-rotor platforms.  Notably, it exhibits distinct temporal and modal discrepancies between UAV and satellite imagery, introducing substantial challenges in CVGL.

\textbf{Modifications for Scale Estimation.}
 To support our experiments, we refined the original datasets as follows:

\textit{DenseUAV+ (Satellite Expansion):} The original DenseUAV dataset only provides pre-cropped satellite image patches with fixed sizes, preventing the validation of scale-adaptive cropping. To resolve this, we downloaded spatially continuous satellite maps from Google Earth, ensuring the satellite gallery covers the entire flight envelope for arbitrary cropping (See Fig.~\ref{fig:dataset_overview}a). Since the original dataset provides the UAV's relative altitude above the takeoff point, we directly use this value as the ground-truth altitude. In our experiments, we employ the official training set to train the CVGL model and utilize the official test subset comprising 2,331 images for evaluation.

\textit{UAV-VisLoc+ (SfM Refinement \& Cleaning).}  The original UAV-VisLoc dataset poses two major challenges: (i) it provides only absolute altitude, whereas visual scale estimation requires relative altitude to the ground; and (ii) the ground truth poses contain inherent sensor noise. To address these issues, we utilized the advanced photogrammetry software \textit{Agisoft Metashape}\footnote{\url{https://www.agisoft.com/}}. Treating the originally provided intrinsics and noisy poses as initial values, we performed structure from Motion on the image sequences. This process refined the camera poses and generated corresponding Digital Surface Model (DSM) (See Fig. \ref{fig:dataset_overview}b). To mitigate the impact of local terrain relief and accidental measurement errors, the ground reference is defined as the mean value of the lowest 10\% of the DSM within a 100m radius of the FOV center. Based on this ground reference, we can precisely calculate the relative altitude for every frame. Furthermore, to ensure rigorous evaluation, we filtered problematic sub-regions: Region 07 and Region 10 were excluded due to insufficient image quantity ($<150$ images); Region 09 was removed due to discontinuous satellite map coverage; and Region 11 was discarded due to extreme terrain relief. Consequently, Regions 01-06 and 08 were selected as the reliable testbed for our experiments. Additionally, UAV images dominated by water surfaces (e.g., rivers) were filtered out, as their lack of texture features hinders effective CVGL. After this data cleaning process, a total of 4,638 images are retained for evaluation.

\subsubsection{Implementation Details}
\label{subsubsec:Details}

\textbf{Scale Estimation Settings.}
 For SV detection, we employ the RTMDet model \cite{lyu2022rtmdet} trained on the VSAI Dataset \cite{VSAI}. During inference, we set the confidence threshold $\tau_{conf}$ to 0.5 and the minimum anchor count threshold $N_{min}$ to 5; For the geometric modeling of SV, we employ empirical metric priors: length $\mu_L = 4.4$ m, width $\mu_W = 1.9$ m, and height $H_{car} = 1.6$ m.
 
\textbf{CVGL Settings.}
Follow \cite{ye2025exploring}, we adopted  CAMP \cite{wu2024camp} method for CVGL and  trained the model with the DenseUAV training set.
During training, data augmentations are applied, including random rotation, grid dropout, and color jitter.
Training is conducted for 3 epochs using the AdamW optimizer, where the initial learning rate is set to $1\times10^{-3}$ and is scheduled by a cosine learning-rate scheduler; the batch size is set to 24.
Both training and inference are performed with an input resolution of $384\times384$. During CVGL process, the UAV query is constructed by center-cropping the largest square region from the UAV image (due to the square input requirement) and resizing it to $384\times384$.
Given the image size, UAV altitude $H$, and the effective focal length $f$, the ground field-of-view range $S_{\mathrm{FOV}}$ is computed, within which satellite patches are cropped from the satellite map using a sliding window with a stride of 50\% of the patch size.
Feature embeddings are extracted by CAMP for the UAV query and all satellite patches, and the center of the patch achieving the maximum similarity is taken as the predicted location.
For evaluation, the ground-truth location is obtained by projecting the UAV camera center (from the UAV pose) onto the ground plane.
A prediction is regarded as correct if the ground-plane distance between the predicted and ground-truth locations is smaller than $\sqrt{2}$ times the UAV FoV radius; otherwise, it is regarded as a failure.
Accordingly, the localization success rate (SR) under this criterion is reported as the CVGL metric. All experiments are conducted on a single NVIDIA GeForce RTX 4090 GPU.

\subsubsection{Sensitivity to Scale Mismatch}
\label{subsubsec:scale_sensitivity}

We firstly design an experiment to analyze how different levels of scale discrepancy affect CVGL performance.
Specifically, for each UAV query, we set a series of adjusted relative altitudes $\tilde{H}=H(1+\delta)$ around the ground truth, where $\delta$ denotes the relative error ratio.
Each $\tilde{H}$ determines the expected ground coverage area, which is used for scale-adaptive satellite cropping and CVGL.
\begin{figure*}[!h]
    \centering
    \begin{subfigure}[t]{0.49\textwidth}
        \centering
        \includegraphics[width=\textwidth]{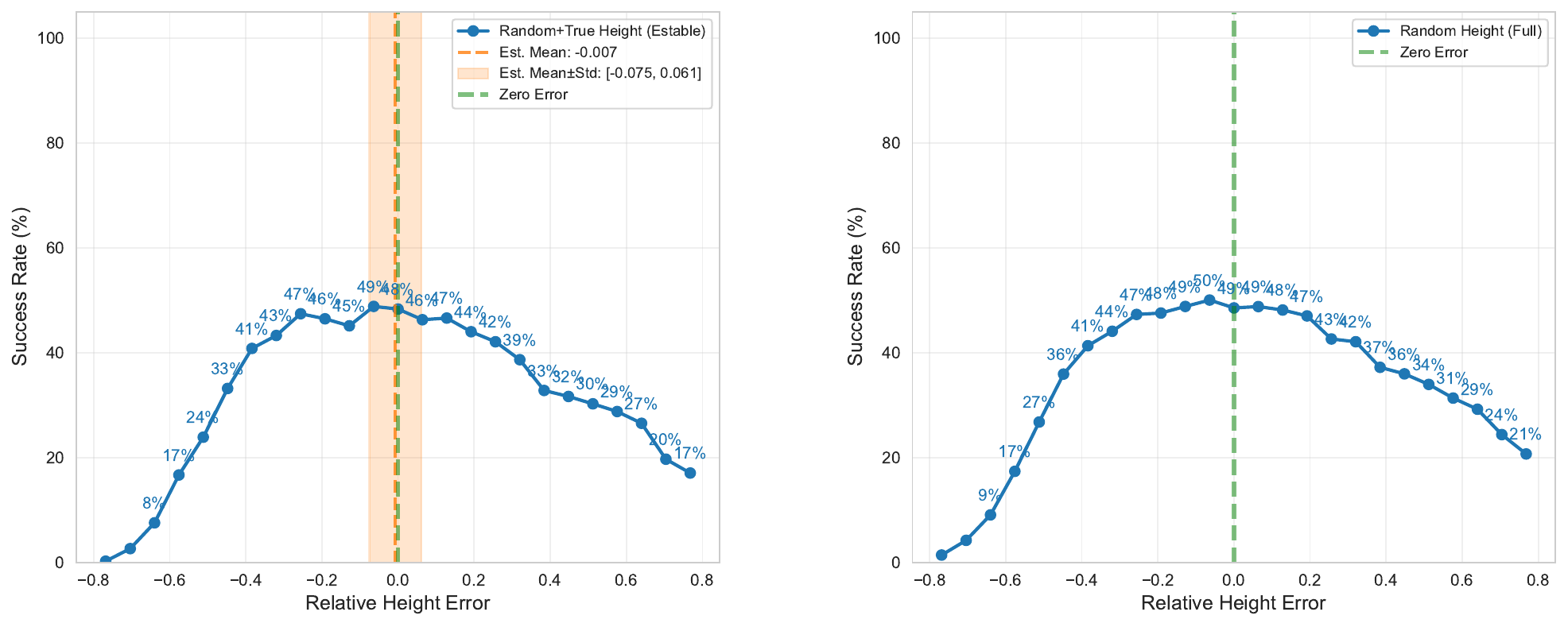}
        \caption{DenseUAV+.}
        \label{fig:scale_curve_denseuav}
    \end{subfigure}
    \hfill
    \begin{subfigure}[t]{0.49\textwidth}
        \centering
        \includegraphics[width=\textwidth]{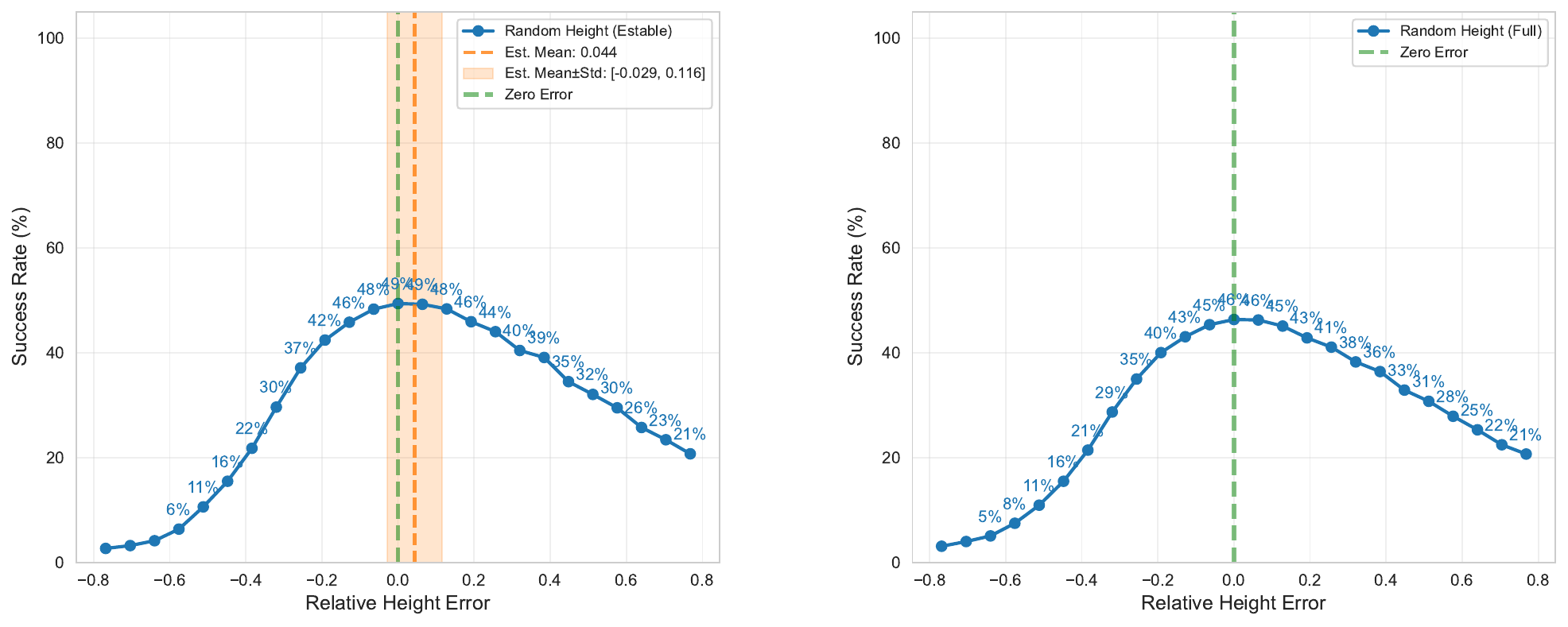}
        \caption{UAV-VisLoc+.}
        \label{fig:scale_curve_uvl}
    \end{subfigure}

    \caption{Sensitivity to scale mismatch.
    Success Rate (SR, \%) under different relative altitude ratios $\delta$ in $\tilde{H}=H(1+\delta)$ (equivalently, relative scale mismatch).
    Each subfigure contains two plots: \textbf{left} reports results on queries where our scale estimator is applicable (with sufficient semantic anchors, e.g., $\tau_{conf}\!\ge\!0.5$ and $N\!\ge\!5$), while \textbf{right} reports results on \emph{all} queries (including those without valid scale estimates).
    The \textcolor{green}{green} vertical line indicates the ground-truth scale ($\delta=0$).
    The \textcolor{orange}{orange} band denotes the distribution of our estimated scale errors, visualized as $\mu\pm\sigma$ of $\delta$ (computed on the estimable subset), showing that most estimates fall into the stable regime where CVGL performance is less sensitive to scale.}
    \label{fig:scale_curve}
\end{figure*}

\textbf{Results.}
Fig.~\ref{fig:scale_curve} reports the localization success rate (SR) under different $\delta$.
The performance is highest when the scale is close to the ground truth (i.e., $|\delta|\leq 10\%$), where the UAV image and the satellite crop cover similar physical areas.
As the scale mismatch increases, SR decreases consistently, indicating that scale discrepancy directly affects cross-view geo-localization.
Moreover, two curves are reported for each dataset: the left plot evaluates queries where scale estimation is applicable (with sufficient semantic anchors), while the right plot includes all queries.
The orange band (mean $\pm$ std of the estimated $\delta$) shows that the estimated scale errors are concentrated around $\delta=0$ and largely fall within the stable regime.


\subsubsection{Scale (Altitude) Estimation Accuracy}
\label{subsubsec:altitude_accuracy}

We further evaluate the proposed method for scale estimation on DenseUAV+ and UAV-VisLoc+.
The quantitative results are summarized in Table~\ref{tab:altitude_res}. In our setting, the satellite crop size is proportional to the UAV altitude under fixed camera intrinsics.
Therefore, the relative altitude error is equivalent to the relative scale error:
\begin{equation}
\frac{|\tilde{H}-H|}{H} \approx \frac{|\tilde{s}-s|}{s}.
\end{equation}
Since altitude is a common and intuitive metric in UAV applications, we report the Mean Absolute Percentage Error (MAPE) of altitude as the evaluation metric.

\textbf{Results and Analysis.}
As shown in Table~\ref{tab:altitude_res}, our method achieves stable accuracy on both datasets, with an overall error of 4.4\% on UAV-VisLoc+ and 2.9\% on DenseUAV+.
The used-image ratio is 50.5\% on UAV-VisLoc+ and 33.7\% on DenseUAV+, indicating that our method can provide scale estimation for a substantial fraction of test images.
We also observe that some UAV-VisLoc+ regions have relatively low used ratios (e.g., R05--R06), mainly because these scenes are more suburban with fewer vehicles (see Fig.~\ref{fig:dataset_overview}b).
To ensure reliability, we only perform scale estimation on images with sufficient semantic anchors, i.e., detections satisfying $\tau_{conf}\geq 0.5$ and $N\geq 5$.
This conservative design ensures that scale estimation is well-constrained whenever it is applied; otherwise, we prefer skipping the image rather than producing unstable estimates.

In addition, DenseUAV+ yields slightly better accuracy than UAV-VisLoc+.
We attribute this mainly to the larger apparent scale of vehicles under low-altitude imaging: vehicles occupy more pixels in the image, which reduces the relative impact of detector localization noise and pixel-level quantization.
Meanwhile, larger object projections provide a stronger metric constraint (i.e., a longer measurement baseline in the image plane), making the scale estimation less sensitive to small localization errors.


\subsubsection{Scale-Adaptive CVGL with Estimated Scale}
\label{subsubsec:retrieval_with_est_scale}

We further evaluate the complete pipeline by performing scale-adaptive CVGL using the estimated scale.
We compare two settings: (i) using the ground-truth altitude for satellite cropping, and (ii) using the estimated altitude from our method.
As shown in Table~\ref{tab:cvgl_gt_vs_est}, the localization success rate under the estimated scale is close to the ground-truth setting on both DenseUAV+ and UAV-VisLoc+ (within about 0.3\% and 1.3\%, respectively).

This observation is also consistent with the sensitivity analysis in Fig.~\ref{fig:scale_curve}.
Since our estimation errors in Table~\ref{tab:altitude_res} are within the stable range (approximately $\pm 10\%$), the CVGL performance remains close to the ground-truth scale setting.


\begin{figure*}[!h]
    \centering
    \includegraphics[width=\textwidth]{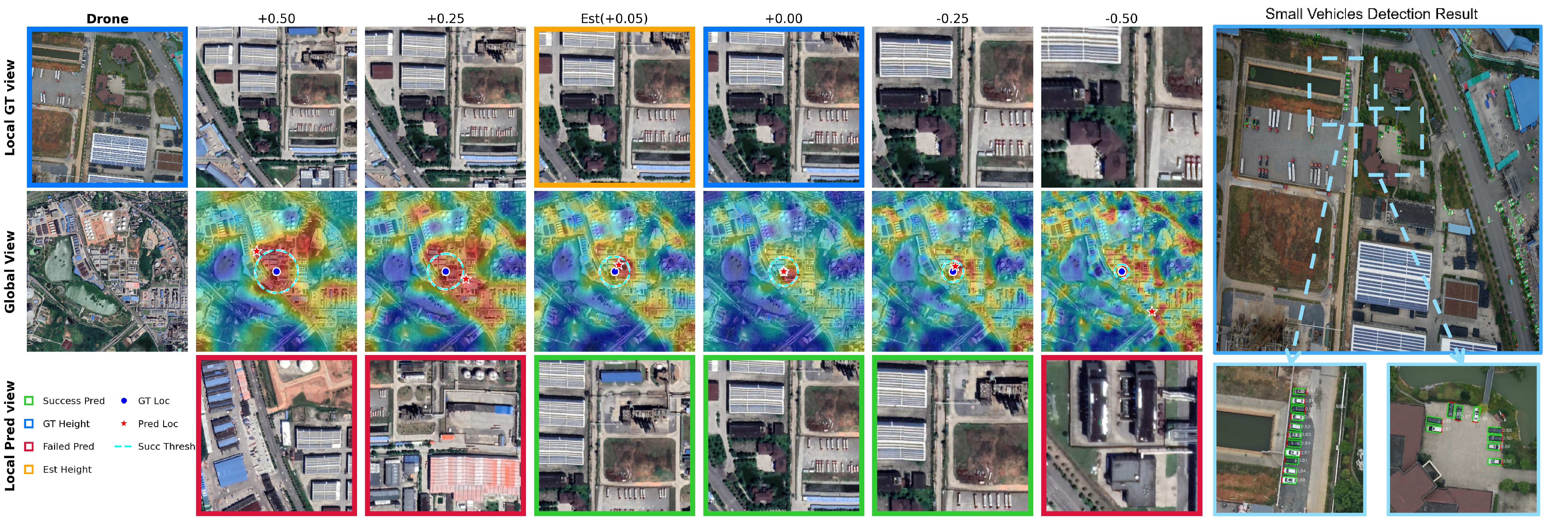} 
    \caption{Visualization of scale mismatch in scale-adaptive CVGL
    We use one UAV-VisLoc+ query and vary the relative altitude ratio $\delta$ in $\tilde{H}=H(1+\delta)$.
    Top: local GT view, showing the UAV query (left) and the GT satellite crops under different $\delta$.
    Middle: global search view, showing the satellite search region (left) and the corresponding similarity heatmaps over the full region.
    Bottom: retrieved satellite patches and final localization outcomes (green/red: success/failure), with detected vehicles used as semantic anchors shown on the right.
    When $\delta=0$ (scale matched), the similarity surface exhibits a clear single dominant peak near the GT location.
    Under scale mismatch (e.g., $\delta=\pm0.25,\pm0.50$), the similarity surface becomes multi-modal and localization easily drifts to spurious peaks.
    Our estimated scale (Est, corresponding to a small relative error ratio, e.g., $\delta\approx +0.05$) recovers reliable localization.}
    \label{fig:scale_visualization}
\end{figure*}
To better visualize the effect of scale mismatch and the benefit of using our estimated scale, Fig.~\ref{fig:scale_visualization} presents a CVGL case study on UAV-VisLoc+.
When the adjusted altitude $\tilde{H}=H(1+\delta)$ matches the ground-truth scale ($\delta=0$), the similarity heatmap forms a clear single dominant peak around the GT location, leading to correct localization.
As $|\delta|$ increases, the similarity heatmap becomes multi-modal and the prediction is prone to drift to incorrect peaks, resulting in failures.
In contrast, our estimated scale yields a small relative error ratio, which effectively mitigates scale mismatch and enables robust CVGL.

Finally, besides scale estimation for CVGL, our method also provides a practical altitude estimate.
This is valuable when UAV metadata is missing or unreliable, and in GNSS-denied scenarios where vision-based altitude initialization is needed.

\begin{table}[h]
    \centering
    \caption{UAV flight altitude estimation performance. True Alt. represents the ground truth relative altitude range; Used denotes the number of images with sufficient semantic anchors ($\tau_{conf} \ge 0.5, N \ge 5$);  Mean Err. indicates the Mean Absolute Percentage Error (MAPE) between estimated and true altitude.}
    \label{tab:altitude_res}
    \scalebox{0.8}{
    \begin{tabular}{l l c c c c c c}
        \toprule
        \textbf{Dataset} & \textbf{Region} & \textbf{\makecell{True Alt. \\ Range (m)}} & \textbf{\makecell{Total \\ Images}} & \textbf{Used} & \textbf{\makecell{Used \\ Ratio}} & \textbf{\makecell{Avg. \\ Cars}} & \textbf{\makecell{Mean \\ Err.}} \\
        \midrule
        \multirow{7}{*}{UAV-VisLoc+} & R01 & 382--452 & 690 & 532 & 77.1\% & 106.4 & 4.3\% \\
                                     & R02 & 350--404 & 818 & 221 & 27.0\% & 34.2  & 5.5\% \\
                                     & R03 & 454--462 & 768 & 515 & 67.1\% & 70.4  & 3.7\% \\
                                     & R04 & 535--551 & 738 & 446 & 60.4\% & 41.1  & 3.7\% \\
                                     & R05 & 377--416 & 473 & 112 & 23.7\% & 24.0  & 5.2\% \\
                                     & R06 & 325--364 & 344 & 72  & 20.9\% & 14.1  & 4.8\% \\
                                     & R08 & 539--595 & 807 & 442 & 54.8\% & 15.3  & 3.3\% \\
        \midrule
        \textbf{UAV-VisLoc+} & \textbf{Overall} & 325--595 & \textbf{4638} & \textbf{2340} & \textbf{50.5\%} & 43.6 & 4.4\% \\
        \midrule
        \textbf{DenseUAV+}   & \textbf{All}  & 80--100 & \textbf{2331} & \textbf{786} & \textbf{33.7\%} & 10.0 & 2.9\% \\
        \bottomrule
    \end{tabular}}
\end{table}

\begin{table}[t]
    \centering
    \caption{Scale-adaptive CVGL performance (SR, \%) using the estimated altitude (Est) versus ground-truth altitude (GT), together with altitude statistics and estimation accuracy. MAPE denotes the Mean Absolute Percentage Error of the estimated altitude.}
    \label{tab:cvgl_gt_vs_est}
    \scalebox{0.85}{
    \begin{tabular}{l l c c c c}
        \toprule
        \textbf{Dataset} & \textbf{Region} &
        \textbf{\makecell{True Alt.\\Range (m)}} &
        \textbf{\makecell{Used\\Ratio}} &
        \textbf{\makecell{Mean Err.\\(MAPE)}} &
        \textbf{\makecell{SR (\%)\\Est / GT}}
 \\
        \midrule
        \multirow{8}{*}{UAV-VisLoc+}
            & R01 & 382--452 & 77.1\% & 4.3\% & 38.0 / \textbf{38.2} \\
            & R02 & 350--404 & 27.0\% & 5.5\% & 53.8 / \textbf{56.6} \\
            & R03 & 454--462 & 67.1\% & 3.7\% & 50.4 / \textbf{52.9} \\
            & R04 & 535--551 & 60.4\% & 3.7\% & 63.2 / \textbf{63.9} \\
            & R05 & 377--416 & 23.7\% & 5.2\% & 71.4 / \textbf{73.2} \\
            & R06 & 325--364 & 20.9\% & 4.8\% & \textbf{65.3} / \textbf{65.3} \\
            & R08 & 539--595 & 54.8\% & 3.3\% & 29.9 / \textbf{31.1} \\
            \cmidrule{2-6}
            & \textbf{Overall} & 325--595 & \textbf{50.5\%} & 4.4\% & 53.1 / \textbf{54.4} \\
        \midrule
        DenseUAV+ & \textbf{All} & 80--100 & \textbf{33.7\%} & 2.9\% & 48.0 / \textbf{48.3} \\
        \bottomrule
    \end{tabular}}
\end{table}

\subsubsection{Comparison with monocular depth estimation.}
\begin{figure*}[t]
    \centering
    \includegraphics[width=\textwidth]{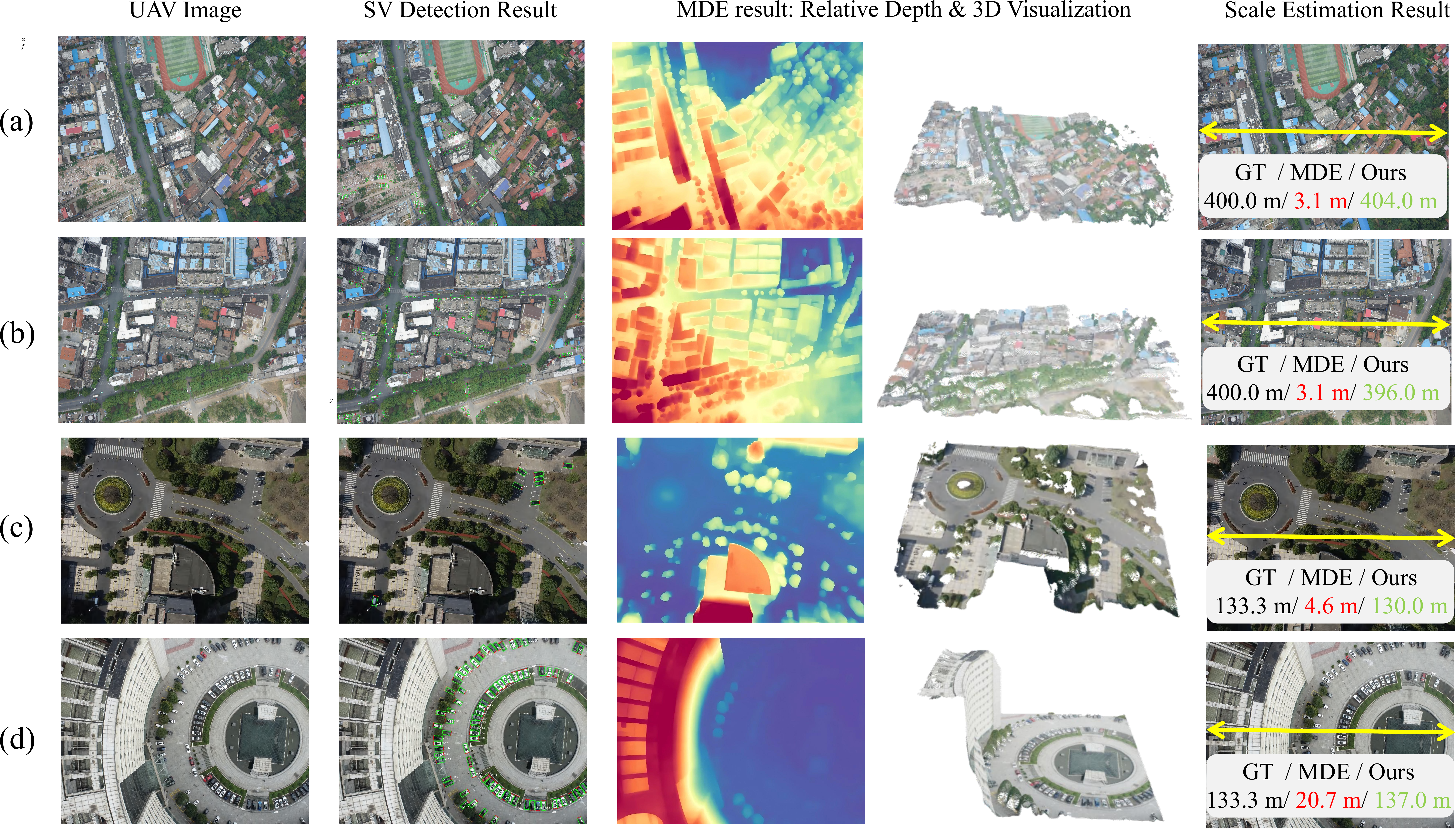}
    \caption{Qualitative comparison with monocular depth estimation. Rows (a)--(b): UAV-VisLoc Dataset; rows (c)--(d): DenseUAV Dataset. 
    Each row shows (left to right) the UAV image, vehicle detections, Depth Anything V3 relative depth/3D visualization, and the estimated ground distance.
    GT: ground truth; MDE: Depth Anything v3; Ours: our method.}
    \label{fig:mde_vs_ours}
\end{figure*}

We further compare our scale recovery with a recent state-of-the-art monocular depth estimation (MDE) model, Depth Anything V3 \cite{depth_anything_v2}.
As shown in Fig.~\ref{fig:mde_vs_ours}, although the predicted \emph{relative} depth can be visually plausible, Depth Anything V3 often fails to produce reliable \emph{metric} (absolute) scale on UAV imagery, leading to severely underestimated distances.
A plausible explanation is the pronounced domain gap: most monocular depth models are trained and evaluated primarily on ground-view data, whereas UAV scenarios involve much larger scale variation due to altitude changes, different viewpoints, and more complex scene layouts, which makes metric-scale generalization across domains difficult.
In contrast, our method enforces explicit geometric constraints and exploits vehicle size priors together with platform-available calibration/pose information, yielding stable and accurate scale recovery across scenes and flight conditions.

\subsection{Ablation Studies}
In this section, we conduct ablation studies to validate the key design choices of our framework and to assess its robustness under different settings.
Following the experimental protocol in Sec.~\ref{subsec:exp_cvgl}, we first analyze the sensitivity of the recovered scale to potential errors in the geometric inputs (i.e., pitch $\theta$ and focal length $f$).
We then compare our decoupled geometric modeling with a naive baseline that directly maps detection-box edges to vehicle dimensions, and finally investigate the influence of hyperparameters as well as the generalizability across different oriented object detectors.

\textbf{Sensitivity to errors in $\theta$ and $f$.}
The pitch angle $\theta$ and focal length $f$ are key parameters in our geometric modeling and directly affect the recovered scale.
We therefore analyze how their systematic biases propagate to the final estimation.
According to Eq.~(\ref{equ:scale_estimate}), the average spatial resolution is
$r=\hat{s}/|\sin\theta|$.
Assuming a small pitch bias $\Delta\theta$ (in radians), a first-order approximation yields
\begin{equation}
\frac{\Delta r}{r}
\approx
\frac{\Delta \hat{s}}{\hat{s}}
-
\cot(\theta)\,\Delta\theta .
\end{equation}
This indicates that the relative error of $r$ is approximately proportional to $\Delta\theta$, scaled by $\cot(\theta)$.

From Eq.~(\ref{equ:alpha}) and Eq.~(\ref{equ:gamma}), the viewing angles $(\alpha,\gamma)$ depend on $f$ through
$\alpha=\arctan\!\big((u-c_x)/f\big)$ and $\gamma=\arctan\!\big((v-c_y)/f\big)$. A first-order perturbation with respect to $f$ gives
\begin{equation}
\Delta\alpha \approx 
-\frac{u-c_x}{f^2+(u-c_x)^2}\,\Delta f,
\qquad
\Delta\gamma \approx
-\frac{v-c_y}{f^2+(v-c_y)^2}\,\Delta f .
\label{eq:sens_f_general}
\end{equation}
Equivalently, using $\tan\alpha=(u-c_x)/f$ and $\tan\gamma=(v-c_y)/f$, Eq.~(\ref{eq:sens_f_general}) can be written as
\begin{equation}
\Delta\alpha \approx -\frac{\sin\alpha\cos\alpha}{f}\,\Delta f,
\qquad
\Delta\gamma \approx -\frac{\sin\gamma\cos\gamma}{f}\,\Delta f,
\end{equation}
showing that the angle perturbations are still first-order in $\Delta f$, with a sensitivity factor varying with the viewing angles.
Since $(L_{eff},W_{eff})$ and the instance scales are smooth functions of $(\alpha,\gamma)$, the induced scale error is first-order in $(\Delta\alpha,\Delta\gamma)$ and thus approximately linear in $\Delta f$. In practice, both $\Delta\theta$ and $\Delta f$ are typically small.
The pitch angle is provided by the onboard IMU with high attitude accuracy, and $f$ is obtained via offline camera calibration with low reprojection error.
Therefore, their impact on scale recovery is limited.

\textbf{Naive Baseline vs. Decouple Modeling.}
To verify the necessity of the proposed decouple modeling, we compare our full model with a naive baseline. In the naive baseline, we directly treat the long and short edges of the detection box as the vehicle length and width. We do not decouple geometric terms and we do not consider stereo-like effects. As shown in Table~\ref{tab:ablation_naive}, the naive baseline leads to much larger errors on both datasets. On DenseUAV+, the MAPE increases from 2.9\% to 8.1\%. This is mainly because DenseUAV+ contains many low-altitude cases (80--100\,m), where perspective distortion and box bias are stronger. On UAV-VisLoc+, the average MAPE increases from 4.4\% to 5.9\%. These results indicate that decouple modeling is important to reduce systematic bias from perspective effects and imperfect detection boxes, and it is needed for more accurate scale recovery.

\begin{table}[t]
    \centering
    \caption{Comparison between our full model and a naive baseline. MAPE denotes the Mean Absolute Percentage Error of the estimated altitude. The default setting is indicated in gray.}
    \label{tab:ablation_naive}
    \resizebox{\columnwidth}{!}{
    \begin{tabular}{l c c}
    \toprule
    Method & DenseUAV+ MAPE (\%) & UAV-VisLoc+ (Avg) MAPE (\%) \\
    \midrule
    \rowcolor{gray!20} Full model (Default) & 2.9 & 4.4 \\
    Naive (box edges as vehicle) & 8.1 (\,$\uparrow$5.2) & 5.9 (\,$\uparrow$1.5) \\
    \bottomrule
    \end{tabular}
    }
\end{table}

\textbf{Hyperparameter Sensitivity.}
We investigate the detection confidence threshold \(\tau_{conf}\) and minimum anchor count \(N_{min}\). As shown in Table~\ref{tab:ablation}, a clear trade-off exists between data availability and estimation precision. Relaxed thresholds (e.g., \(\tau_{conf}=0.2\)) maximize coverage but introduce substantial geometric noise from false positives, leading to unacceptable errors ($>14\%$). Conversely, increasing \(\tau_{conf}\) beyond 0.5 yields diminishing returns in accuracy (e.g., 2.9\% \(\to\) 2.2\% on DenseUAV+) while causing a sharp decline in usage ratio. Similarly, while a higher \(N_{min}\) enhances statistical stability, it excludes valid scenes with sparse targets. Consequently, we adopt \(\tau_{conf}=0.5\) and \(N_{min}=5\) as the default settings to strike a balance between accuracy and usage ratio.

\textbf{Generalizability across Detectors.}
We further assess universality by integrating diverse oriented object detectors, including RTMDet, Rotated-FCOS \cite{tian2019fcos}, R\(^3\)Det \cite{yang2021r3det}, and ROI-Transformer \cite{ding2019learning}. As summarized in Table~\ref{tab:ablation}, our framework enables valid scale recovery across all models. Under the default settings, RTMDet yields the most robust performance. Notably, model sensitivity varies; ROI-Transformer exhibits suboptimal performance at \(\tau_{conf}=0.5\) due to false positives but achieves competitive accuracy (3.6\%) when \(\tau_{conf}\) is adjusted to 0.9, albeit with a lower usage ratio. This confirms that while our geometric framework is model-agnostic, optimal hyperparameters should be calibrated according to the specific confidence distribution of the underlying detector.

\begin{table}[t]
    \centering
    \caption{Ablation studies on hyperparameters and detection models. Ratio indicates the usage ratio (availability), and MAPE denotes the Mean Absolute Percentage Error of the estimated altitude. The default setting is indicated in gray.}
    \label{tab:ablation}
    \resizebox{\columnwidth}{!}{
    \begin{tabular}{l|cc|cc|cc}
    \toprule
    \multirow{2}{*}{Method/Settings} & \multicolumn{2}{c|}{Parameters} & \multicolumn{2}{c|}{DenseUAV+} & \multicolumn{2}{c}{UAV-VisLoc+ (Avg)} \\
    \cline{2-7}
     & \(\tau_{conf}\) & \(N_{min}\) & Ratio (\%) & MAPE (\%) & Ratio (\%) & MAPE (\%) \\
    \midrule
    \multicolumn{7}{l}{\textit{Different \(\tau_{conf}\)}} \\
    RTMDet & 0.2 & 5 & 49.0 & 15.7 & 78.5 & 14.6 \\
    RTMDet & 0.3 & 5 & 39.4 & 5.8 & 61.2 & 7.7 \\
    \rowcolor{gray!20} RTMDet (Default) & 0.5 & 5 & 33.7 & 2.9 & 50.4 & 4.4 \\
    RTMDet & 0.7 & 5 & 29.9 & 2.3 & 38.6 & 3.6 \\
    RTMDet & 0.8 & 5 & 26.1 & 2.2 & 25.4 & 3.4 \\
    \midrule
    \multicolumn{7}{l}{\textit{Different \(N_{min}\)}} \\
    RTMDet & 0.5 & 2 & 43.7 & 3.5 & 64.1 & 5.0 \\
    \rowcolor{gray!20} RTMDet (Default) & 0.5 & 5 & 33.7 & 2.9 & 50.4 & 4.4 \\
    RTMDet & 0.5 & 8 & 21.9 & 2.5 & 41.5 & 3.5 \\
    RTMDet & 0.5 & 10 & 15.1 & 2.5 & 38.1 & 3.2 \\
    \midrule
    \multicolumn{7}{l}{\textit{Different Detectors}} \\
    \rowcolor{gray!20} RTMDet (Default) & 0.5 & 5 & 33.7 & 2.9 & 50.4 & 4.4 \\
    Rotated-FCOS & 0.5 & 5 & 29.5 & 3.7 & 42.1 & 3.7 \\
    R\(^3\)Det-OC & 0.5 & 5 & 32.9 & 3.8 & 48.6 & 4.9 \\
    ROI-Trans & 0.5 & 5 & 39.9 & 11.9 & 57.6 & 10.0 \\
    ROI-Trans & 0.8 & 5 & 33.8 & 5.5 & 42.1 & 5.1 \\
    ROI-Trans & 0.9 & 5 & 31.7 & 3.6 & 37.9 & 4.6 \\
    \bottomrule
    \end{tabular}
    }
\end{table}

\subsection{Application: Absolute Scale Recovery for Orthophotos}
\label{subsec:exp_ortho}

Beyond individual UAV images, our method extends to recovering the absolute scale for 3D reconstructed scenes (e.g., orthophotos). In many real-world engineering and urban planning scenarios, 3D models generated from monocular videos or historical data inherently lack absolute metric scale due to the absence of camera intrinsics or ground control points (GCPs). This scale ambiguity poses a fundamental challenge: designers cannot accurately assess whether a planned structure fits into the existing environment. Unlike traditional solutions that rely on labor-intensive on-site GCP measurements or manual alignment with reference satellite imagery, our approach offers a purely visual solution to recover the global metric scale, serving as a critical prerequisite for metric-sensitive downstream tasks.

\textbf{Experimental Setup.}
We utilize high-resolution orthophotos generated from the \textbf{UAV-VisLoc+} dataset to validate our method. Since orthophotos represent a nadir-projected planar view, the stereoscopic effects caused by object height can be negligible compared to oblique imagery. Therefore, we disregard the vehicle height ($H_{car}$) and treat detected bounding boxes as direct 2D planar projections to estimate the GSD.

\textbf{Scale Recovery Performance.}
The quantitative results are summarized in Table \ref{tab:ortho_scale}. The proposed framework achieves a remarkably low average relative error of \textbf{2.6\%} across all tested regions. Notably, regions with dense semantic anchors, such as Region 01 ($N=13,066$), achieve relative errors as low as 2.4\%. Even in Region 06, which presents a challenging scenario with sparse anchors ($N=198$), the error remains within a reasonable range (7.2\%). These results confirm that our method can reliably recover the global metric scale from pure visual data, providing precise scale initialization for unscaled reconstruction results.

\begin{figure*}[t]
    \centering
    \includegraphics[width=1.0\linewidth]{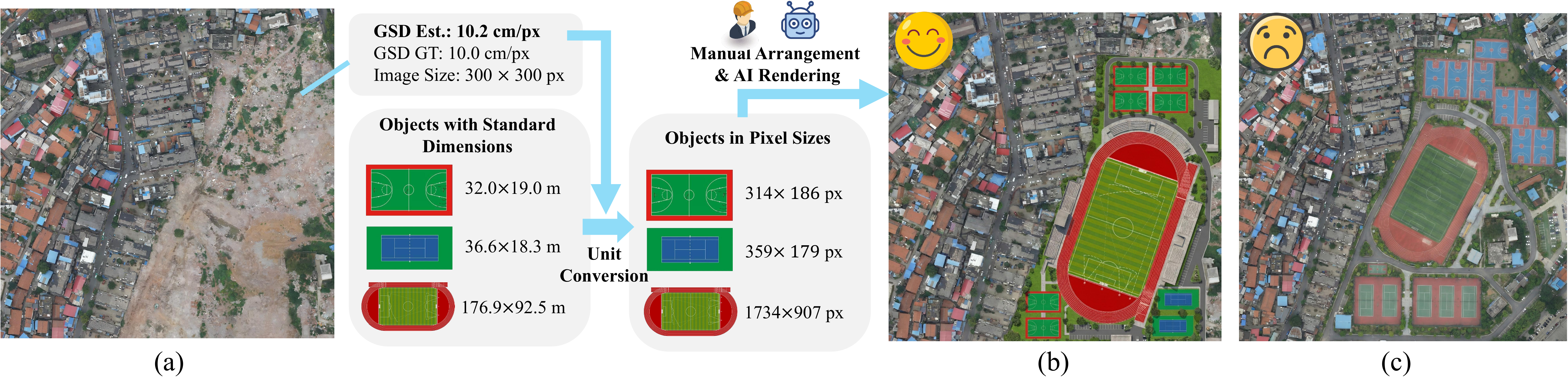} 
    \caption{Metric-Aware Generative Urban Planning. We simulate a site planning task on an unscaled orthophoto crop. (a) The input crop from Region 01 representing a vacant lot. (b) Ours (Metric-Aware Generation): Using the GSD ($10.2$ cm/px) estimated by our method, standard sports facilities were strictly scaled to pixel dimensions, manually laid out, and then rendered using Nano Banana Pro. The result exhibits correct physical proportions consistent with the environment. (c) Direct Generation: The result rendered directly by Nano Banana Pro without metric constraints. The generated facilities show significant scale deviation (e.g., the stadium is unrealistically large).}
    \label{fig:rendering}
\end{figure*}

\textbf{Metric-Aware Generative Urban Planning.}
To demonstrate the practical necessity of our method, we design a simulation experiment for \emph{planning on an unscaled map}, as illustrated in Fig.~\ref{fig:rendering}. The objective is to design a sports complex on a vacant lot within a reconstructed orthophoto (Fig.~\ref{fig:rendering}a). Without scale information, translating real-world facility dimensions into the image domain is ambiguous and often infeasible.
Our framework first estimates the GSD of the unscaled map as $10.2$~cm/px (Ground Truth: $10.0$~cm/px). Leveraging the recovered metric, we build a \textit{Scale-Aware} pipeline: we convert standard facilities (e.g., a $28 \times 15$~m basketball court) into pixel space according to the estimated GSD, place them within the vacant lot to form a plausible semantic layout, and use the resulting layout as geometric constraints for a generative model. Concretely, we prompt \textit{Nano Banana Pro} \footnote{\url{https://nanobanapro.com/}} to \emph{render the surrounding context} given that the facilities have already been placed with approximately correct metric scale (i.e., ``design a sports complex on the vacant lot; the courts and track have been placed in near-real scale---please render the surrounding region accordingly''). As shown in Fig.~\ref{fig:rendering}b, the generated facilities are physically consistent with nearby roads and buildings, and the overall plan exhibits coherent proportions.
For comparison, we perform a \textit{Direct Generation} baseline using \textit{Nano Banana Pro} without any metric constraints or pre-placed layout (Fig.~\ref{fig:rendering}c). The prompt only specifies the desired facilities and requests scale consistency with the scene (i.e., ``design a sports complex on the vacant lot; place a track field and several basketball/badminton courts, and keep the scale consistent with other regions''). The resulting structures exhibit severe scale distortion, where the stadium becomes disproportionately large compared to nearby streets, rendering the plan unusable. This comparison demonstrates that our recovered scale makes unscaled imagery usable for engineering applications.

\begin{table}[h]
    \centering
    \caption{Absolute scale recovery results on UAV VisLoc+ orthophotos. The Ground Sampling Distance (GSD) is reported in cm/pixel. All errors are absolute values.}
    \label{tab:ortho_scale}
    \scalebox{0.95}{
    \begin{tabular}{l c c c c c}
        \toprule
        \textbf{Region} & \textbf{Cars Used} & \textbf{GT GSD} & \textbf{Est. GSD} & \textbf{Abs. Err.} & \textbf{Abs. Err.} \\
        & ($N$) & (cm/px) & (cm/px) & (cm) & (\%) \\
        \midrule
        01 & 13,066 & 10.0 & 10.2 & 0.2 & 2.4\% \\
        02 & 1,562  & 10.0 & 10.0 & 0.0 & 0.0\% \\
        03 & 7,404  & 12.0 & 11.8 & 0.2 & 1.7\% \\
        04 & 3,679  & 14.0 & 13.7 & 0.3 & 1.8\% \\
        05 & 475    & 13.0 & 12.9 & 0.1 & 1.1\% \\
        06 & 198    & 8.0  & 7.4  & 0.6 & 7.2\% \\
        08 & 1,450  & 14.0 & 13.5 & 0.5 & 3.8\% \\
        \midrule
        \textbf{Mean} & \textbf{3,976} & \textbf{11.6} & \textbf{11.4} & \textbf{0.3} & \textbf{2.6\%} \\
        \bottomrule
    \end{tabular}}
\end{table}

\section{Conclusion}
In this paper, we explore the feasibility of recovering the absolute scale for UAV-to-satellite cross-view geo-localization (CVGL) from semantic anchors with metric priors.
By leveraging detected small vehicles and a semantic--geometry formulation, reliable scale estimates are obtained and integrated into a scale-adaptive CVGL pipeline.
Extensive experiments on DenseUAV+ and UAV-VisLoc+ demonstrate that accurate scale estimation can be achieved (e.g., 2.9\% and 4.4\% MAPE on the two datasets), while the resulting scale-adaptive CVGL attains localization performance close to the ground-truth setting with only minor success rate (SR) drops.
Beyond scale-adaptive CVGL, the proposed estimator can also provide practical altitude initialization when UAV metadata is missing or unreliable, and can be used to recover the metric scale of unscaled aerial reconstructions (e.g., orthophotos) for downstream metric-aware applications.

While the proposed framework shows promising results, its effectiveness is currently strongest in man-made areas where small vehicles are sufficiently present and can be detected reliably, and it may be less applicable to natural scenes with sparse semantic anchors. Moreover, for high-altitude or low-resolution imagery, vehicles may occupy too few pixels to be detected robustly, which can reduce the reliability of the estimated scale.
Nevertheless, this work provides evidence that semantic anchors can serve as practical cues for metric initialization in UAV-to-satellite CVGL.
Future research can further improve robustness and coverage by incorporating additional semantic cues with weak but useful metric regularities (e.g., coarse building footprints or road-width statistics), and by developing multi-object coupling strategies that fuse heterogeneous targets for scale estimation.

\bibliographystyle{IEEEtran}
\bibliography{main}












\section{Biography Section}

\vspace{-40pt}
\begin{IEEEbiography}[{\includegraphics[width=1in,height=1.25in,clip,keepaspectratio]{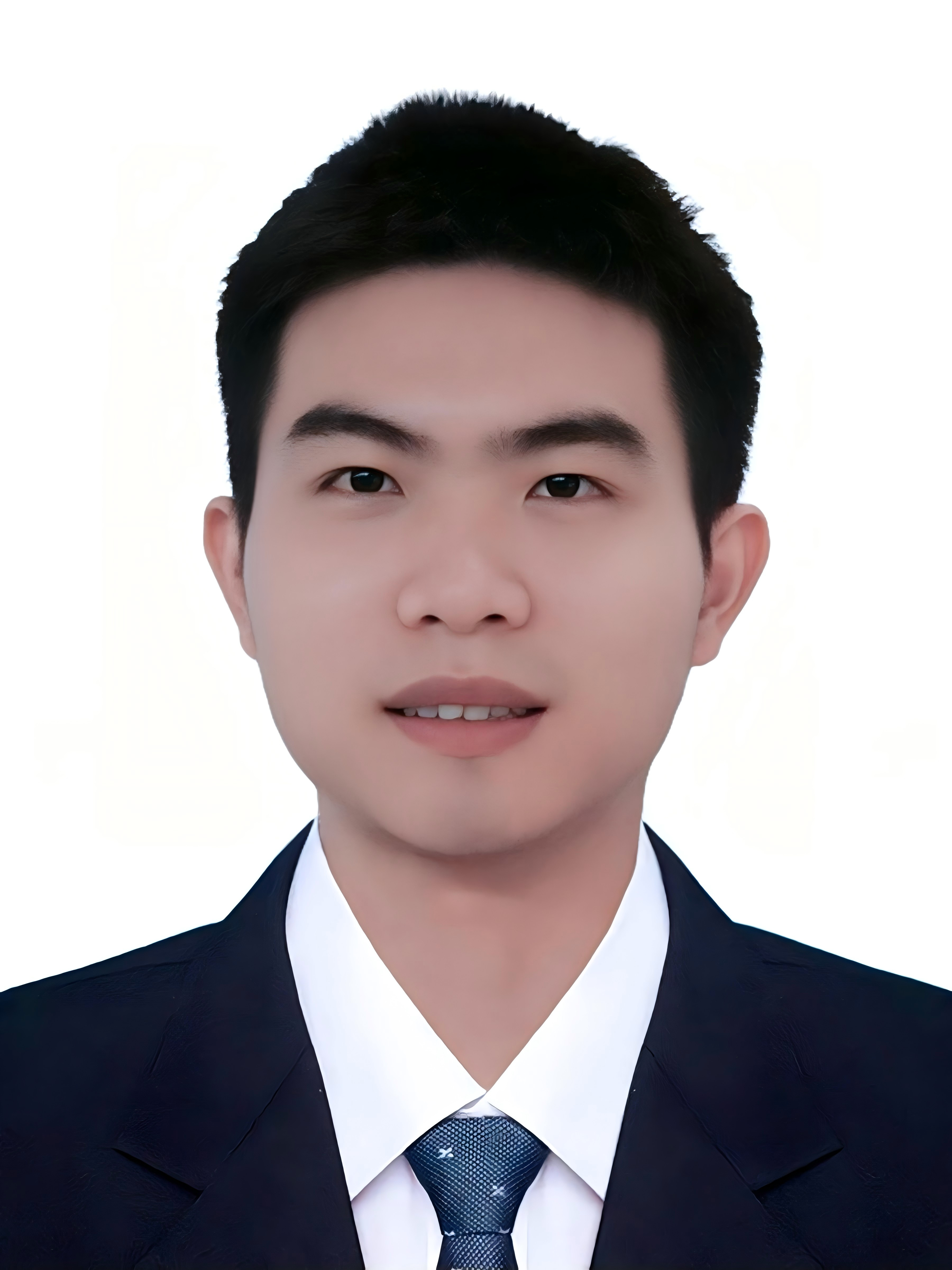}}]{Yibin Ye}
received the B.E. degree in National University of Defense Technology (NUDT), Changsha, China, in 2022. He is currently pursuing the Ph.D. degree with the College of Aerospace Science and Engineering, NUDT. His current research interests include multimodal image matching and UAV visual navigation.
\end{IEEEbiography}
\vspace{-31pt}
\begin{IEEEbiography}
[{\includegraphics[width=1in,height=1.25in,clip,keepaspectratio]{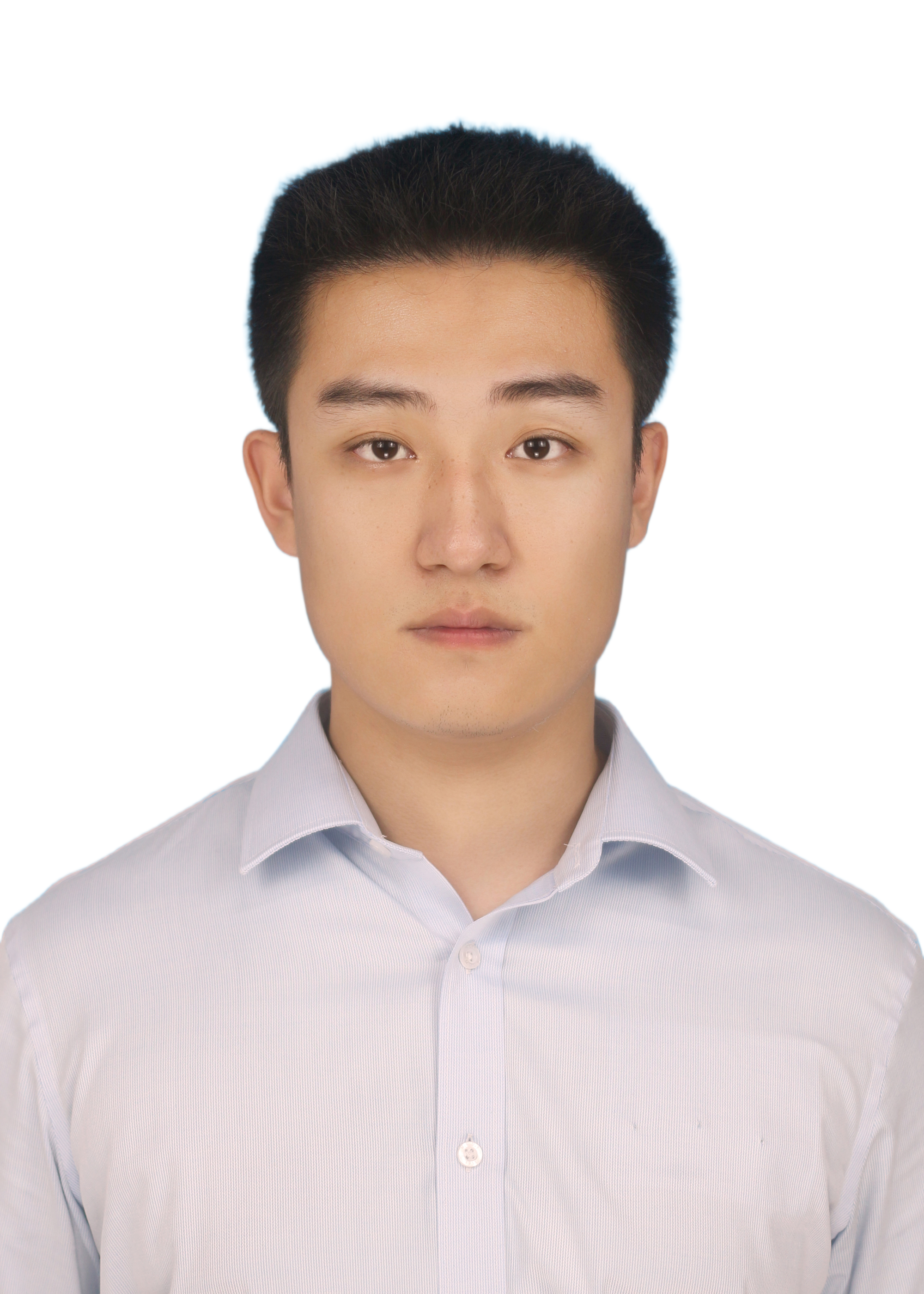}}]{Shuo Chen}
received the B.E. degree in National University of Defense Technology (NUDT), Changsha, China, in 2024. He is currently pursuing the master's degree with the College of Aerospace Science and Engineering, NUDT. His current research interests include cross-view geo-localization.
\end{IEEEbiography}
\vspace{-31pt}
\begin{IEEEbiography}
[{\includegraphics[width=1in,height=1.25in,clip,keepaspectratio]{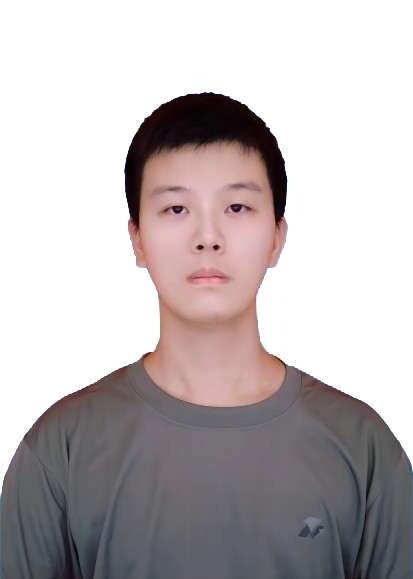}}]{Kun Wang}
received the B.E. and M.E. degrees from Wuhan University (WHU), Wuhan, China, in 2018 and 2020.
He received the Ph.D. degree in Aeronautical and Astronautical Science and Technology from National University of Defense Technology (NUDT), Changsha, China, in 2025.
His research interests include oriented object detection, knowledge distillation, and photogrammetry.
\end{IEEEbiography}
\vspace{-31pt}
\begin{IEEEbiography}
[{\includegraphics[width=1in,height=1.25in,clip,keepaspectratio]{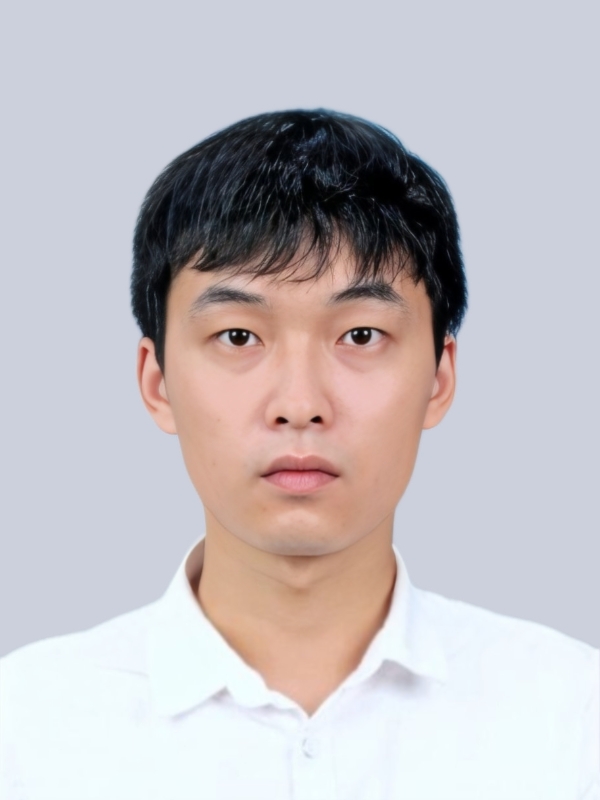}}]{Xiaokai Song}
received his B.E. and M.E. degrees from the School of Automation, Wuhan University of Technology, Wuhan, Hubei Province, China, in 2019 and 2022, respectively. He is currently pursuing his Ph.D. degree at the National University of Defense Technology, Changsha, Hunan Province, China. His research interests include UAV visual navigation and simultaneous localization and mapping (SLAM).
\end{IEEEbiography}
\vspace{-31pt}
 \begin{IEEEbiography}
[{\includegraphics[width=1in,height=1.25in,clip,keepaspectratio]{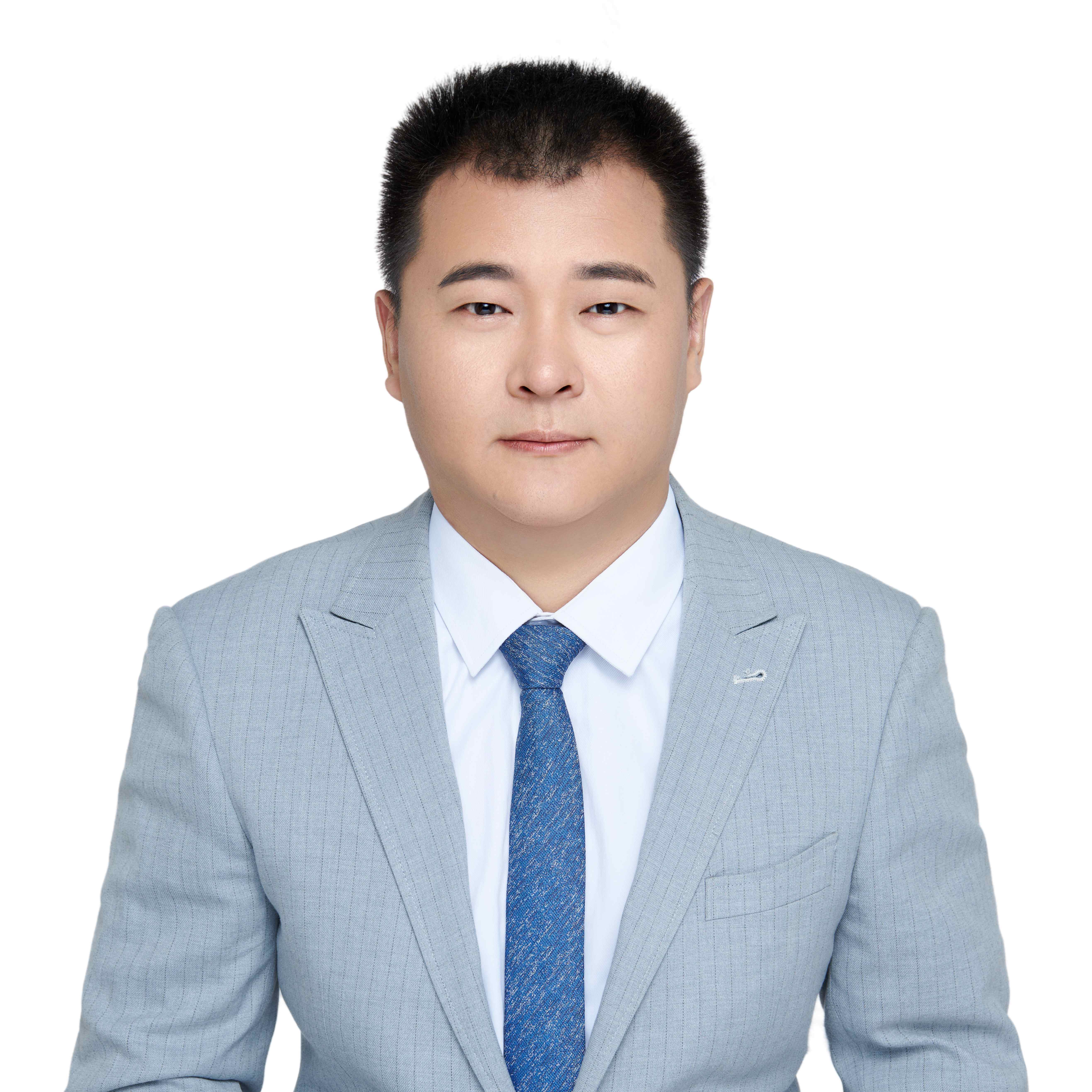}}]{Jisheng Dang}
received the Ph.D. degree from Sun Yat-sen University, China, in 2025, advised by prof. Jianhuang Lai and prof. Huicheng Zheng.
He worked as a research fellow at the NExT++ laboratory of the National University of Singapore  advised by prof. Tat-Seng Chua. He is now a tenured associate professor at the School of Information Science and Engineering, Lanzhou University. His research interests include multimodal learning, video understanding, and embodied  intelligence. 
He has published several papers as the first author in major journals and conferences including IEEE TIP/TNNLS/TITS/IJCAI/AAAI. He served as a reviewer at some major journals and conferences like IEEE TPAMI, ICML, NIPS, ICLR, IEEE TIP, CVPR, IJCAI, ACM MM, AAAI, IEEE TMM.
\end{IEEEbiography}
\vspace{-31pt}
\begin{IEEEbiography}
[{\includegraphics[width=1in,height=1.25in,clip,keepaspectratio]{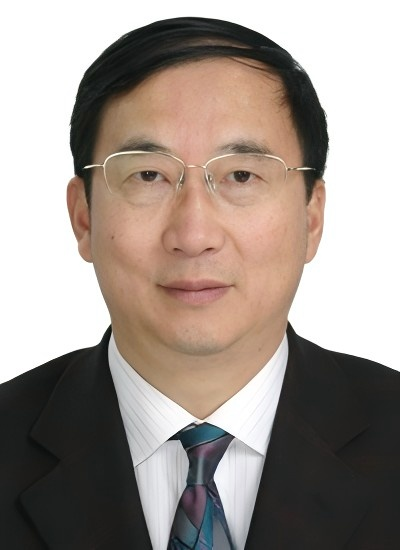}}]{Qifeng Yu} 
received the B.S. degree from Northwestern Polytechnic University, Xi’an, China, in 1981, the M.S. degree from the National University of Defense Technology, Changsha, China, in 1984, and the Ph.D. degree from Bremen University, Bremen, Germany, in 1996. 
He is a Professor with the National University of Defense Technology and a member of the Chinese Academy of Sciences. 
He has authored three books and published over 100 articles. 
His main research fields include image measurement, vision navigation, and close-range photogrammetry. 
\end{IEEEbiography}
\vspace{-31pt}
 \begin{IEEEbiography}
[{\includegraphics[width=1in,height=1.25in,clip,keepaspectratio]{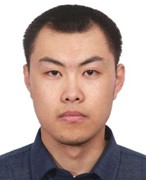}}]{Xichao Teng}
received the B.S. degree from Beihang University, Beijing, China, in 2012 and the M.S. and Ph.D. degrees from National University of Defense Technology, Changsha, China, in 2015 and 2019, respectively. He is currently an instructor in National University of Defense Technology. His research interests include computer vision, remote sensing, and machine learning.
\end{IEEEbiography}
\vspace{-31pt}
\begin{IEEEbiography}
[{\includegraphics[width=1in,height=1.25in,clip,keepaspectratio]{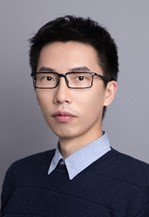}}]{Zhang Li}
received the B.S. degree in communication engineering from National University of Defense Technology, China, in 2008 and the Ph.D. degree in biomedical engineering from Delft University of Technology, Netherlands, in 2015. 
He is now an associate professor with National University of Defense Technology. His research interests include multi-modal (biomedical) image analysis, particularly image registration. He is also interested in the applications of deep learning in remote sensing and computer aided diagnosis system. He initiated the ACDC challenge in conjunction with ISBI 2019 and MICCAI 2020. He also co-organized COMPAY Workshop in MICCAI 2019 and 2021. He is a Junior Editor of Acta Mechanica Sinica, Member of Chinese Space Remote Sensing Professional Committee, Member of Chinese Experimental Mechanics Professional Committee.
\end{IEEEbiography}





\end{document}